# SciDoc2Diagrammer-MAF: Towards Generation of Scientific Diagrams from Documents guided by Multi-Aspect Feedback Refinement

**Ishani Mondal,**[1] **Zongxia Li,**[1] **Yufang Hou,**[2] **Anandhavelu Natarajan,**[3]
**Aparna Garimella,**[3] **Jordan Boyd-Graber**[1]

[1] University of Maryland, College Park,   [2] IBM Research Dublin  , [3] Adobe Research
{imondal@umd.edu, zli12321@umd.edu, YHou@ie.ibm.com,
anandhavelu@gmail.com, garimell@adobe.com, ying@umd.edu}

## Abstract

Automating the creation of scientific diagrams from academic papers can streamline the development of tutorials, presentations, and posters. Current text-to-image models struggle with generating accurate and visually appealing diagrams from long-context inputs. We propose SciDoc2Diagram, a task that extracts relevant information from scientific papers and generates diagrams, along with a benchmarking dataset, SciDoc2DiagramBench. We develop a multi-step pipeline SciDoc2Diagrammer that generates diagrams based on user intent using intermediate code generation. We observed that initial diagram drafts are often incomplete or unfaithful to the source, requiring SciDoc2Diagrammer-Multi-Aspect-Feedback (MAF), a refinement strategy that significantly enhances factual correctness and visual appeal and outperforms existing models on both automatic and human judgement.

## 1 Introduction

Researchers often need to transform detailed studies into compelling diagrams for presentations, going beyond the usual figures in their papers to include summaries, methodologies, and dynamic data visualizations (Chapman et al., 2014).[1] Automation of diagram generation from the multimodal components of academic papers—such as text, figures, tables, and plots—could enable the creation of accurate and appealing visual aids for scientific tutorials and presentations (Mondal et al., 2024; Fu et al., 2021; Sun et al., 2021), which can help researchers and educators to improve productivity and enhancing the overall quality of their presentations. We introduce a new task SciDoc2Diagram where the input comprises of the content of academic papers and user's intent and the output is a scientific diagram crafted from the content of papers and aligned with user's intent, which we call "Extrapolated-Diagram".

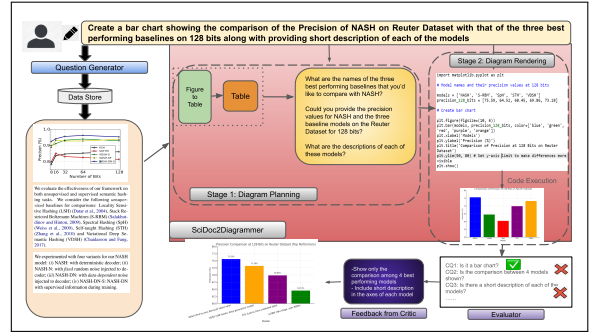

Figure 1: An example of Diagram Generation and Refinement using SciDoc2Diagrammer-MAF with input as a paper and user-defined intent.

Consider a scenario where a busy scientist wants to create figures from papers within a tight deadline, with the specific goal of generating a bar chart as shown in the Figure 1. To achieve this quickly, recent advancements in text-to-image (T2I) generation models like Stable Diffusion (Rombach et al., 2022a) and DALL-E (Ramesh et al., 2021, 2022a) offer promising solutions. However, these models face limitations in generating factually correct and visually appealing scientific figures. They primarily produce raster graphics at low resolutions, which is not suitable for scientific figures requiring precision and valid text (Hsu et al., 2021). Belouadi et al. (2023) introduced a scientific diagram generation method based on only user's intent. However, creation of informative diagrams using long-content papers as an additional input remains unexplored.

How effective are the current state-of-the-art Large Language models (LLMs) or Visual Language Models (VLMs) at generating scientific diagrams? Despite their potential, a significant knowledge gap exists: there is currently no standard benchmark to assess their performance in this area. To address this gap, we introduce the Sci-

---

[1] According to Fu et al. (2021), approximately 39.2% of figures in ACL slides are creatively derived from the original papers, taking forms like flowcharts, summary tables, and plots/charts that visually interpret the paper's content.

Doc2DiagramBench dataset that features 1080 diagrams from 89 *ACL papers, sourced from existing slides in ACL Anthology (Section 3). This dataset supports researchers in automating scientific diagram creation for presentations.

We evaluate state-of-the-art LLMs (Jiang et al., 2023; Touvron et al., 2023; OpenAI, 2023; Abdin et al., 2024) to extract relevant information from documents pertinent to the user-specified intent. Specifically, we use a multi-step pipelined method, SciDoc2Diagrammer that first extracts relevant data from academic papers corresponding to intent of the user in creating the diagram, generates an intermediate code that renders the final diagram (Section 4.1) as shown in Figure 1. However, the first-draft images generated by these models suffer from a range of errors: hallucination, incompleteness and unreadability due to their incapabilities of parsing long-prompts (Lee et al., 2024a).

Therefore, we propose a sequential refinement approach SciDoc2Diagrammer-MAF based on the intial draft generated by SciDoc2Diagrammer. The new approach is guided by several critic models to provide targeted feedback (multiple aspects such as *Completeness*, *Faithfulness*, and *Look and Feel*) in a step-by-step hierarchical manner on the output to further enhance the quality of output (Section 4.2). Our multi-aspect feedback refinement approach significantly improves diagram quality, as confirmed by both automatic and human evaluations, demonstrating notable enhancements over existing methods (Section 6). Complex flowcharts and tables derived from single or multiple documents show notable enhancements in faithfulness and completeness when refined sequentially. It is particularly advantageous when iteratively synthesizing information from multiple sources to enhance clarity and precision.[2]

## 2 Task Formulation of SciDoc2Diagram

Our goal is to automatically generate scientific diagrams from academic papers tailored to user specifications. We process the input document $D = \{T, F, \tau, \pi\}$, which includes body text $T$, figures $F$, tables $\tau$, and plots $\pi$, guided by a user intent $I$ (e.g., "*Create a bar chart showing the comparison...*" in Figure 18). The output, an Extrapolated-Diagram $\delta$, visually represents content not explicitly illustrated in the paper.

---

[2]Code and data available at https://github.com/Ishani-Mondal/SciDoc2DiagramGeneration.

| Conversion Type | # Diagrams |
|---|---|
| Extrapolated-Flowchart | 320 |
| Extrapolated-Results | 290 |
| Extrapolated-Architecture | 270 |
| Extrapolated-Summary | 200 |
| **Total** | **1080** |

Table 1: Statistics of SciDoc2DiagramBench Dataset which includes 1080 diagrams and the distribution highlights the diversity of diagrams.

## 3 SciDoc2DiagramBench: A New Dataset

Creating Extrapolated-Diagram from source papers is an important yet challenging task. However, no current dataset evaluates state-of-the-art models on this task. To advance research, we have created SciDoc2DiagramBench dataset, repurposing DOC2PPT (Fu et al., 2021) (scientific papers and corresponding slides from ACL Anthology) and TutorialBank (Fabbri et al., 2018) (tutorials from *ACL conferences).[3] Our dataset includes the following three parts and the overall statistics of different diagram types are in Table 1 with examples in Appendix 15.

**SciDoc2DiagramBench-Gold.** Our initial goal was to explore what are the types of Extrapolated-Diagram usually created by humans while making presentations. We hired an expert annotator from Upwork with more than five years of authoring NLP-related papers and creating presentations. The annotator had two tasks, first to identify slides containing Extrapolated-Diagram (excludes slides that used external information not found in the paper), second to manually write an intent for the corresponding diagram creation (13, 12). Through this annotation process, we collect 265 gold-standard Extrapolated-Diagram from 34 papers where each instance comprises of a tuple of <papername, intent of diagram, gold diagram>. The first three authors manually checked the intents, and only instances that were unanimously deemed reasonable were retained. After analyzing these diagrams, 32% were flowcharts illustrating methodology (Extrapolated-Flowchart), 33% were results/plots (**Extrapolated-Results**), 25% were summary of related works/contributions (**Extrapolated-Summary**) and 10% were related to clarifying existing architecture (**Extrapolated-Architecture**).

---

[3]Data Contamination issues are discussed in Appendix 14

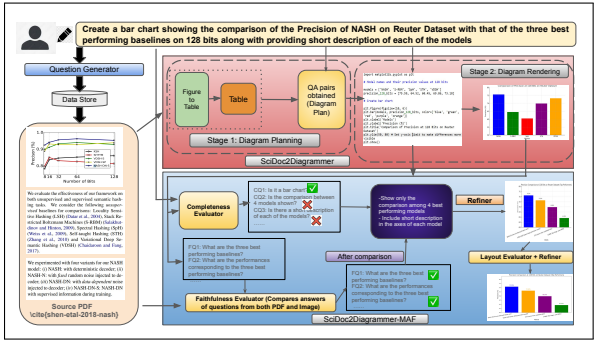

Figure 2: Outlines the procedure for generating diagrams from academic papers based on specific user intents, followed by refinement of each critic models.

**SciDoc2DiagramBench-Extended.** Using insights from SciDoc2DiagramBench-Gold on how presenters transform paper content into diagrams, we used GPT-4 to generate new diagram intents for this subset of data (Table 14). We also provide guidelines for creating Extrapolated-Diagram that blend text and visuals from academic papers. Each intent, reviewed by the first author, aims to guide the creation of practical diagrams. Each instance includes <paperName, intent> but lacks a corresponding human-created gold diagram.

**SciMultiDoc2DiagramBench-Gold.** To evaluate the task on another challenging set, we examined scientific ACL tutorials from TutorialBank (Fabbri et al., 2018), focusing on slides where creators used content from multiple papers to design their diagrams. An expert annotator from Upwork identified these slides with Extrapolated-Diagrams. The first author then manually wrote intents for these tutorial slides, with each instance documented as <paperNames, intent, diagram>.

## 4 Methodology

The SciDoc2Diagram task comprises of generating initial first-draft generation of diagrams from documents using base models (Section 4.1), followed by refinement of diagrams using critic modules (Section 4.2).

### 4.1 SciDoc2Diagrammer

We introduce SciDoc2Diagrammer, a two-stage pipeline for generating "*Extrapolated-Diagrams*" from paper content (Upper portion of Figure 2). In the first *Diagram Planning* stage, a planner VLM takes the PDF (decomposed into sections and figures/tables) and user's intent and generates a diagram plan. Inspired by the success of extracting information using question-generation and retrieving answers from relevant snippets as done by Krithara et al. (2023); Lee et al. (2023); Qiu et al. (2018); Auer et al. (2023), we also synthesize diagram plan by first generating clarifying questions based on the user's intent, and then obtain answers of those questions from relevant portions of the paper to come up with question-answer pairs, which we refer to as diagram plan. In the second stage, inspired by the success of LLMs in generating code from natural language instructions (Ugare et al., 2024; Maddigan and Susnjak, 2023), we use LLMs to take the diagram plan and user intent as inputs to generate an intermediate code and execute the code to finally render the diagram.

**Diagram Planning:** Given an input document $D = \{T, F, \tau, \pi\}$ consisting of body text $T$, figures $F$, tables $\tau$, and plots $\pi$, and user intent $I$, the diagram planning phase includes the following sub-steps: **[1] Intent Classification**: We use an LLM (Few-shot with 4 exemplars) to classify if the intent of the user is related to results or not. For example in Figure 3, the intent is not related to results, but the one in Figure 2 is related to results. **[2] Question Generator**: Using the intent, we ask LLMs to generate clarification questions to create the diagram conditioned on intent. For example, for the intent in Figure 2, we obtain a set of clarification questions. **[3] Retrieve Information:** We retrieve relevant passages from the paper based on each of questions (Lewis et al., 2020). In Figure 2, a plot and two paragraphs from Experimental Section of the source paper are retrieved to answer the generated questions. If any result type figure is relevant to the question, we first extract data from the figures and convert them into tables for simplification of content retrieval as motivated by prior works (Liu et al., 2023a; Masry et al., 2023). **[4] Diagram Plan Synthesis:** After generating questions and extracting relevant content from paper, we extract answers of those questions. Then we formulate a dictionary of QA pairs/diagram plan $Q = \{(q_1, a_1), (q_2, a_2), \ldots, (q_n, a_n)\}$, where $q_i$ are the questions and $a_i$ are the answers.

**Diagram Rendering:** In this phase, we use the user intent $I$ and QA pairs $Q$ from the previous step to generate an intermediate code $\Theta$ that constructs diagram. This intermediate code is represented as Python code, as shown in Figure 2. Finally, we execute the intermediate code $\Theta$ to render

"Extrapolated-diagram" $\delta$, as exemplified by the bar chart at the top-right corner of Figure 2.

### 4.2 SciDoc2Diagrammer-MAF

**Motivation:** To gain a qualitative understanding of the diagrams generated from SciDoc2Diagrammer, we randomly sample 50 diagrams from the test set for manual inspection. There are some frequently occurring errors (Table 10) and we classify them into three broad categories: **Completeness** (essential components are missing or incomplete, affecting the diagram's informativeness), **Faithfulness** (misleading information that do not accurately reflect the source data), and **Layout** (poor visual alignment hindering readability and comprehension). Besides, we also noticed that most of the errors propagate during the transition from our diagram planning to rendering stage; this was due to the inability of LLMs to handle long-context prompts (Nathani et al., 2023; Gani et al., 2024).

Inspired by the self-refinement strategies by Madaan et al. (2023), we wanted to study whether the quality of our diagrams can be enhanced by using generic feedback for a wide range of errors or by applying targeted feedback for specific error categories. For exploring the second strategy, we introduce an iterative refinement framework, depicted in Figure 2 (Lower part), designed to enhance quality of diagram generation by specifically rectifying above-mentioned error types.

**Overview of the Components:** As illustrated in Figure 3, our MAF framework comprises of the following components: a base model (SciDoc2Diagrammer) that generates an initial diagram ($O$) using both intent and source document (Step 1 in 3). It passes through three specialized feedback critic evaluator modules, each focusing on a critical aspect of diagram quality such as completeness ($C_{critic}$), faithfulness ($F_{critic}$), and layout ($L_{critic}$) on $D$, a Refiner module $R$ that generates refined diagram $O'$ based on $O$ and feedback.

**Feedback Modules:** The completeness critic $C_{critic}$ (Algorithm 1) decomposes the diagram's intent $I$ and generates a set of questions $Q$ from the user intent $I$ to determine how well various elements are represented in the diagram (Step 3). It then involves extracting answers from the diagram, scoring each response for adequacy, calculating an average score as the completeness score $C_{\text{Score}}$, and providing textual feedback ($C_{\text{Feedback}}$) to suggest improvements (Step 4).

The Faithfulness critic $F_{critic}$ (Algorithm 2) evaluates how accurately a diagram represents information from its document source. It evaluates the accuracy of a diagram by generating questions, retrieving answers from both the document and the diagram, and scoring their consistency to provide an average score and feedback for improvement. The layout critic $L_{critic}$ (Algorithm 4) evaluates how good is the look and feel of generated diagram, and provides score and feedback for improvement. We use two refinement techniques:

**Summarization-Based Refinement Algorithm (SumMAF)** We adopt summarization-based multi-aspect refinement strategy from Nathani et al. (2023) for diagram refinement. However, in contrast to their method, we use our critic models that act as human proxies (Algorithm 4), and revisit PDF source and user intent repeatedly to validate and refine the diagram $C_{critic}$(Algorithm 1) and $F_{critic}$(Algorithm 2), mimicking the iterative human process of cross-verifying information through continuous questioning and answering.

The first draft is sent to $C_{critic}$ (Algorithm 1) which checks if all necessary components are present (In Figure 3, it checks the completeness of creating a flowchart). At the same time, the faithfulness evaluator $F_{critic}$ (Algorithm 2) ensures the accuracy of diagram by comparing it with source PDF, whether or not all the information in the source figure is true, by generating questions (FQ1, FQ2 in Figure 2) based on the diagram and obtaining answers from both PDF and diagram. Also, the layout evaluator checks the visual arrangement, ensuring good alignment and spacing (Figure 2). All the critic models provide feedback on a single output (Step 3') and sent to the refiner. The refiner model makes adjustments by summarizing the suggested modifications from all critics and generates a new diagram (Step 4'). This is repeated (Steps 3', 4', 5' in Figure 3) until the evaluator models are satisfied or we reach maximum iterations.

**Sequential Refinement Algorithm (SeqMAF):** Algorithm 3 enhances a diagram in a step-by-step hierarchical manner ensuring that each critic is satisfied (See Figure 18). The intuition is that, complex tables or flowcharts benefit from being accurate and detailed in their depiction of processes or relationships, where sequential refinement allows for gradual addition of details, ensuring that each element is correctly represented. Here the diagram is feed into different critic modules one by one, refine the output until each of them are satisfied.

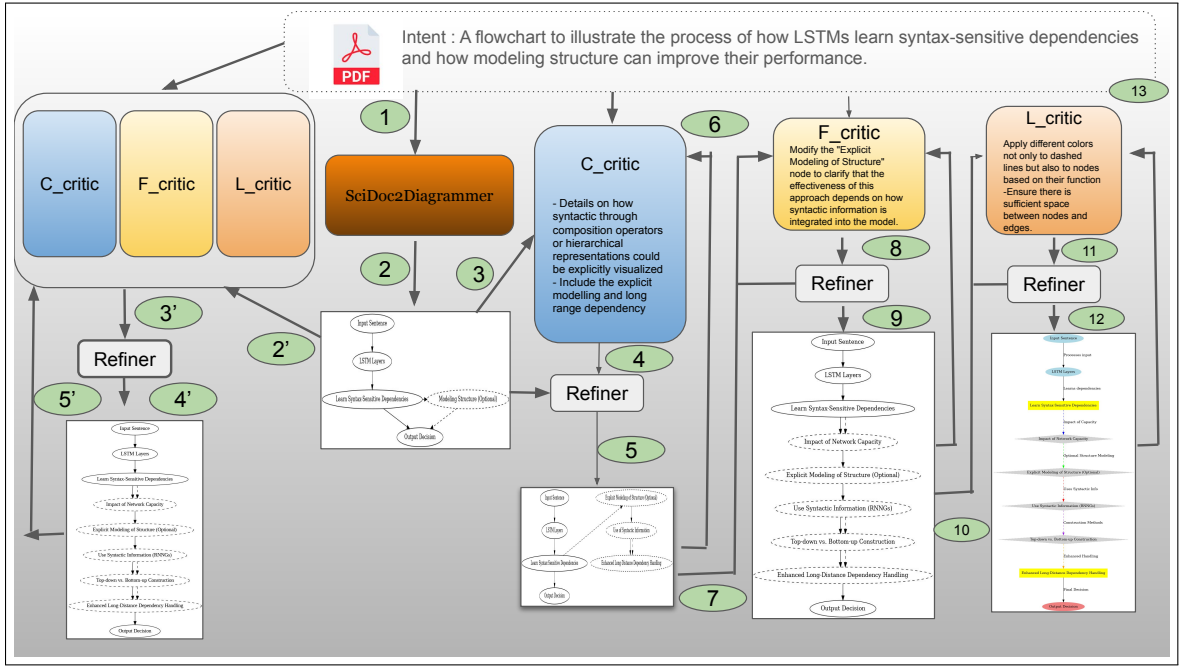

Figure 3: The figure (an example from SciDoc2DiagramBench) depicts SciDoc2Diagrammer-MAF, which refines diagrams based on user intent and source document. Initially, SciDoc2Diagrammer creates a diagram (Step 2), which is refined through three critic modules which assess and provide feedback on necessary components, data accuracy, and visual design. The diagram is repeatedly refined, as shown on the left side of the figure (illustrating SumMAF), where feedback from all critics is integrated at Step 4'. The refinement continues, evaluated at Step 5', until the maximum iterations are reached or the diagram meets the specified standards.

In Figure 3, $C_{critic}$ first provides feedback (detail composition operators or hierarchical representation, explicit modelling and long range dependency step inclusion) on the diagram in Step 4, and the refiner takes the initial figure and refines based on feedback in Step 5, and passed onto $C_{critic}$ to further evaluate and repeat the same steps (4, 5, 6) till maximum iteration is used or the $C_{critic}$ provides a score greater than 4.5 (as stated in Algorithm 1. The same process is repeated by $F_{critic}$ and $L_{critic}$ modules and generates output after Step 12.

## 5 Experimental Setup

**Dataset Preprocessing.** Text on the papers in SciDoc2DiagramBench was extracted through scipdf Parser[4] which has been built on top of the Grobid (GRO, 2008–2024). Figures and captions were extracted through pdffigures2 (Clark and Divvala, 2016). We use 20 randomly selected diagrams from SciDoc2DiagramBench with stratified sampling to ensure that we have four type of diagrams (Table 1) for creating in-context examples in our prompts and the rest of the dataset is used for evaluation.

**Base LMs.** For generating first-draft of diagrams, we experiment with state-of-the-art models to extract table from existing figures during diagram planning (Figure 2): 1) MatCha (Liu et al., 2023b), 2) Chart-to-Text (Kantharaj et al., 2022), 3) UniChart (Masry et al., 2023), 4) DePlot (Liu et al., 2023a) and GPT-4 Vision (GPT-4V) (OpenAI, 2023). Finally, based on our assessment (Figure 20), we concluded that GPT-4V outperformed other models and was the best choice. For diagram planning, we experiment with three open-source LLMs: 1) LLama3 (Huang et al., 2024), 2) Phi-3 (Abdin et al., 2024) Mistral-7b (Jiang et al., 2023) and for diagram rendering we use GPT4-o (OpenAI, 2023).

**Critic Models for SciDoc2Diagrammer-MAF.** Inspired by the success of GPT-4V in making pairwise comparison of evaluation of images using Question-Answering driven approaches (OpenAI, 2023; Zhang et al., 2023) and image aesthetic evaluation (Abe et al., 2024), we adopt GPT-4V model to provide feedback on different aspects of our generated diagrams on various criteria (Prompts can be found in 20, 21, 22).

---
[4]https://github.com/titipata/scipdf_parser

|  | Flowchart | | | Results | | | Architecture | | | Summary | | |
| --- | --- | --- | --- | --- | --- | --- | --- | --- | --- | --- | --- | --- |
| Models (IV) | R | BS | CS | R | BS | CS | R | BS | CS | R | BS | CS |
| DALLE-3 | 0.12 | 0.45 | 0.23 | 0.11 | 0.34 | 0.21 | 0.25 | 0.45 | 0.25 | 0.04 | 0.10 | 0.18 |
| Automatikz | 0.20 | 0.48 | 0.25 | 0.20 | 0.50 | 0.29 | 0.24 | 0.45 | 0.28 | 0.27 | 0.56 | 0.45 |
| SciDoc2Diagram | | | | | | | | | | | | |
| w/ GPT4-o (ZS) | 0.22 | 0.58 | 0.43 | 0.32 | 0.49 | 0.30 | 0.24 | 0.50 | 0.34 | 0.33 | 0.61 | 0.62 |
| w/ GPT4-o (FS) | 0.28 | 0.67 | 0.43 | 0.40 | 0.56 | 0.38 | 0.32 | 0.57 | 0.45 | 0.45 | 0.67 | 0.62 |
| w/ GPT4-o-SR | 0.30 | 0.70 | 0.45 | 0.44 | 0.56 | 0.38 | 0.38 | 0.65 | 0.54 | 0.47 | 0.68 | 0.69 |
| w/ GPT4-o-SumMAF | 0.35 | 0.74 | 0.48 | 0.50 | 0.57 | 0.39 | 0.34 | 0.64 | 0.49 | 0.50 | 0.74 | 0.74 |
| w/ GPT4-o-SeqMAF | 0.39 | 0.79 | 0.53 | 0.49 | 0.49 | 0.36 | 0.37 | 0.58 | 0.49 | 0.45 | 0.67 | 0.62 |

Table 2: Automatic evaluation of models on various diagrams on **SciDoc2DiagramBench-Gold** using ROUGE (R), BERTScore (BS), CLIPScore (CS) with the highest values shown in green, highlighting that flowcharts, architectures are of the best quality after sequential refinement, whereas the other ones after summarization-based refinement.

| Model | Flowchart | | | Summary | | | Architecture | | | Results | | |
| --- | --- | --- | --- | --- | --- | --- | --- | --- | --- | --- | --- | --- |
|  | C | F | L | C | F | L | C | F | L | C | F | L |
| GPT4-o w/o refinement | 3.3 | 3.5 | 4.0 | 3.9 | 4.0 | 4.2 | 2.8 | 4.0 | 4.0 | 3.7 | 4.0 | 4.0 |
| GPT4-o w/ refinement | 3.9 | 4.0 | 4.0 | 4.5 | 4.1 | 4.2 | 3.2 | 4.0 | 4.0 | 3.7 | 4.5 | 3.8 |
| Phi-3 w/o refinement | 3.2 | 4.0 | 4.0 | 3.8 | 4.0 | 4.0 | 2.8 | 4.0 | 4.0 | 3.5 | 3.6 | 3.8 |
| Phi-3 w/ refinement | 3.8 | 4.0 | 4.3 | 4.4 | 4.1 | 4.2 | 3.0 | 4.3 | 4.0 | 3.8 | 4.1 | 3.8 |

Table 3: Comparison of Completeness (C), Faithfulness (F), and Layout (L) assessed by humans for different model variants on **SciDoc2DiagramBench-Extended**, indicating that on the complex set, Flowcharts and Summary tables improve significantly after refinement, but Layout-satisfaction is still comparatively lower.

**Prior Work and Baselines.** We have compared our approaches with existing text-to-image generation methods like DALLE3 (Rombach et al., 2021) and Automatikz (Belouadi et al., 2023). For experimenting with Diagram Planning in SciDoc2Diagrammer, we evaluate using zero-shot and few-shot (with 3 in-context exemplars) versions of LMs as mentioned in Table 8 (Prompts in 15, 16, 18, 19). We use Self-Refine by Madaan et al. (2023) as another baseline (Prompt can be found in 25).

**Evaluation Metrics.** SciDoc2DiagramBench-Gold and SciMultiDoc2DiagramBench-Gold include diagrams explicitly crafted by humans for each document or multiple documents, which can be served as a gold standard, allowing us to benchmark the quality of our generated diagrams. To evaluate our diagrams, we use three automated metrics: **BERTScore** (Zhang et al., 2019), **ROUGE-1** (Lin, 2004), and **CLIPScore** (Hessel et al., 2021).

However, simple token evaluations and neural evaluations like BERTScore and CLIPScore are only measuring string similarity and lacking a more finegrained evaluation framework that correlates more with human preferences (Li et al., 2024c). Thus, we go beyond string similarity scores and include a comprehensive analysis of more finegrained aspects through human evaluation and the GPT4-V Evaluation. We assess **completeness** (measures the degree to which all relevant and necessary information is in the generated figure), **faithfulness** (assesses how well the figure adheres to the facts, data, or specific instructions provided, ensuring that the content is correct and not misleading or misrepresented), and **layout** (measures visual clarity, focusing on how well the elements are structured and arranged) on a scale from 1 to 5.

## 6 Results and Findings

Our main objective is to explore the creation of 'Extrapolated' scientific visuals from documents using zero-shot/few-shot settings of advanced VLMs, and also analyze whether the quality of these visuals without dataset-specific fine-tuning, can be improved using our proposed SciDoc2Diagrammer-MAF. To answer this broad question, we focus our experiments on answering the following research questions as follows:

### 6.1 Main Research Questions and Results

**RQ1: What is the best base LM for generating diagrams before refinement?** GPT4-o is clearly the best-performing base/refiner model on human and automatic judgement. In Table 13, GPT-4o consistently outperforms other models across all categories of diagrams on the SciDoc2DiagramBench.

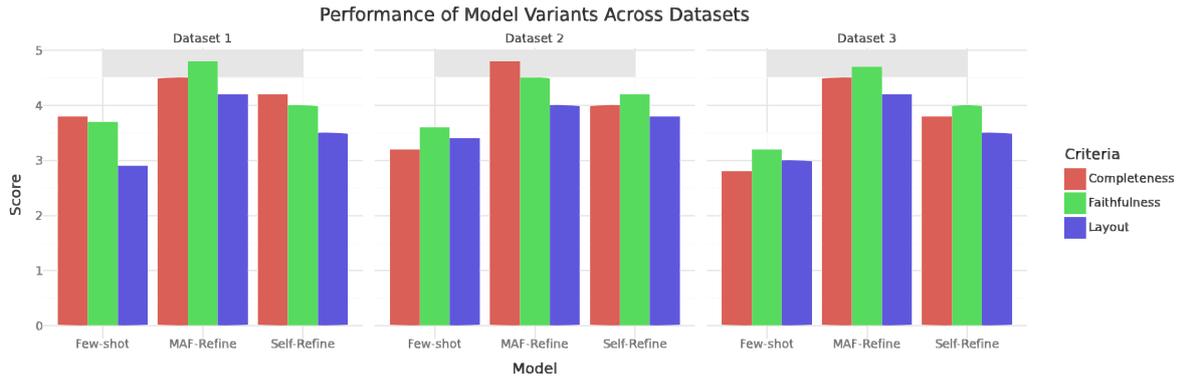

Figure 4: Average Human Rating on Completeness, Faithfulness and Layout on three sub-parts: **SciDoc2DiagramBench-Gold** (left), **SciDoc2DiagramBench-Extended** (middle) and **SciMultiDoc2DiagramBench-Gold** (right), it implies *as the complexity of diagram creation increases, our proposed Multi-Aspect Refinement strategy appears to become more effective.*

| Models (IV) | FChart (R) | FChart (BS) | FChart (CS) | Results (R) | Results (BS) | Results (CS) | Arc (R) | Arc (BS) | Arc (CS) | Summary (R) | Summary (BS) | Summary (CS) |
|---|---|---|---|---|---|---|---|---|---|---|---|---|
| w/ GPT4-o-SeqMAF | 0.39 | 0.79 | 0.53 | 0.49 | 0.49 | 0.49 | 0.36 | 0.37 | 0.58 | 0.45 | 0.67 | 0.62 |
| w/o Completeness-SeqMAF | 0.25↓ | 0.55↓ | 0.35↓ | 0.45 | 0.44 | 0.34↓ | 0.36 | 0.57 | 0.47 | 0.30↓ | 0.50↓ | 0.45↓ |
| w/o Faithfulness-SeqMAF | 0.25↓ | 0.55↓ | 0.35↓ | 0.35↓ | 0.38↓ | 0.25↓ | 0.37 | 0.58 | 0.49 | 0.42 | 0.64 | 0.57 |
| w/o Layout-SeqMAF | 0.33 | 0.74 | 0.52 | 0.49 | 0.49 | 0.49 | 0.36 | 0.54 | 0.46 | 0.41 | 0.62 | 0.59 |

Table 4: Ablation Analysis Results: Here the column names FChart indicates the Extrapolated-Flowchart category of images, Arc indicates the Extrapolated-Architecture category of images, Results denotes the Extrapolated-Results category of images, and Summary denotes the Extrapolated-Summary category of images. Significant drop compared to the first row (greater than 0.1) is shown by ↓.

This superior performance is indicated by the green cells, which show GPT-4o achieving the highest scores in all the automatic metrics. We observe similar trend when we ask annotators to rate on the completeness, faithfulness and layout-satisfaction on 30 randomly selected samples of images from SciDoc2DiagramBench-Gold generated by each model (Figure 19).

**RQ2: How effective is our Scidoc2diagrammer-MAF over SciDoc2Diagrammer?** Based on Table 2, the effectiveness of our refinement strategies on SciDoc2DiagramBench-Gold using GPT4-o with SumMAF and SeqMAF enhancements shows significant improvements over the Zero-shot (ZS) and Few-shot (FS) versions on all the diagram types on automatic evaluation. This finding holds true for SciMultiDoc2DiagramBench-Gold as well, as evident from the results in Table 7.

Drawing inspiration from Ribeiro et al. (2020), we acknowledge that automatic metrics alone are insufficient to fully evaluate model performance, so we also conduct human evaluation. For SciDoc2DiagramBench Gold and SciMultiDoc2DiagramBench-Gold, we had human raters evaluate the quality of diagrams produced by different models and compare these to actual human-created slides, considering the source paper and the diagram's intended purpose for 30 randomly sampled diagrams. In the SciMultiDoc2DiagramBench-Extended, which lacks a gold standard image, only the paper, the intent of the diagram, and the model-generated images were presented for evaluation on 50 randomly sampled images. Two authors and two professional crowdworkers from Upwork assessed the quality of all images, using criteria such as Completeness, Faithfulness, and Layout from 1 to 5. Figure 4 illustrates that *diagrams generated using our refinement strategies compared to Few-shot and Self-Refine strategies consistently receive the highest scores from humans* on all dataset subsets.

**RQ3: What type of refinement works the best across each diagram type based on automatic evaluation?** From Table 2, *SeqMAF shows significant improvement with the highest scores in all metrics compared to both ZS and FS* for flowcharts. SumMAF also improves over ZS and FS, but to a lesser extent than SeqMAF, indicating SeqMAF's superior refinement capability on both completeness and detail accuracy. *For Results diagrams, both SeqMAF and SumMAF show improvements over non-refined versions, with SeqMAF slightly leading in overall performance except for ClipScore.* However, in Architecture and

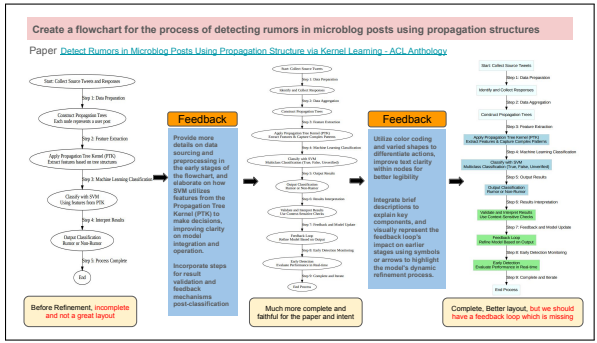

Figure 5: Refinement of a flowchart for detecting rumors from Ma et al. (2017), *highlighting improvements in clarity and completeness after refinement but emphasizing the continued absence of a critical feedback loop.*

Summary diagrams, *SumMAF occasionally outperforms SeqMAF, indicating its strength in contexts requiring synthesized summarization of content*. Since summarization-based refinement offers an integrated approach where multiple facets of feedback are considered simultaneously, this might have been more effective for diagrams requiring balanced presentation of dense information. On SciDoc2DiagramBench-Extended, we explore whether human judgment perceives a significant enhancement in diagram quality across all categories on complex intents. We notice in Table 3, that even though there is a significant improvement of quality post-refinement on human judgement on Faithfulness and Completeness (Flowcharts and tables improve significantly), Layout-satisfaction is the least (which can be attributed to subjectivity and LLM's inability to refine on specific criteria).

**RQ4: Is refinement more effective on simple images or complex images?** Across three datasets, the Multi-Aspect Refinement strategy consistently outperforms the few-shot approach in diagram production (Figure 4). On the simplest dataset, SciDoc2DiagramBench-Gold, MAF-Refine performs significantly better than few-shot approach, with improvements of 0.9 in completeness, 1.1 in faithfulness, and 0.8 in layout. As complexity increases in SciDoc2DiagramBench-Extended, MAF-Refine still performs strongly, showing improvements of 1.4 in completeness, 1.0 in faithfulness, and 0.5 in layout. Even in the most complex dataset, SciMultiDoc2DiagramBench-Extended, MAF-Refine leads with significant improvements, showing that MAF-Refine significantly boosts diagram quality across datasets and maintains its benefits even as complexity rises.

### 6.2 Ablation Analysis

**Impact of individual components in Sci-Doc2Diagrammer.** Table 9 presents the performance impact of different components in the Sci-Doc2Diagrammer model using different metrics. Removing the Question Generation/Answering (QG/QA) component results in significant performance drops across all metrics, highlighting its importance. The absence of the Summarization component also negatively affects the scores but to a lesser extent.

**Impact of Refinements in SciDoc2Diagrammer.** We conduct ablation analysis (Table 4) with SciDoc2Diagrammer-SeqMAF. In particular, we experimented with the removal of each refinement module on the SciDoc2DiagramBench-Gold dataset, and analyzed which component is useful to improve the diagram generation. Completeness removal shows a significant drop in both summary and flowchart metrics, emphasizing the importance of comprehensive overviews and detailed visual representations. **Faithfulness removal** leads to a significant performance drop in results and flowcharts, underlining the necessity of faithful representation for coherent diagram creation. But **layout removal** has the least impact on diagram quality improvement.

### 6.3 Qualitative Analysis

Qualitative analysis in feedback-based refinement (Figure 5) identifies gaps in the initial design, such as missing components or unclear integration of elements. After refinement, while the layout and completeness may improve significantly, issues like the absence of critical feedback loops can persist, leading to a less effective refinement process. This shows that even with visible improvements, missing critical features can detract from the model's intended dynamic refinement functionality.

Besides, taking a closer look at the diagrams generated before and after refinement (we sample the refined image having the highest average score rated by humans), we show in Figure 6, 21, 22 where our algorithm improves. For such qualitative analysis, we consider all four types of extrapolated images. The refined versions of the diagrams appear to show a higher level of detail and precision. For example, in the block diagrams and flowcharts, the GPT4-o with refinement diagrams show clearer layouts and better alignment of elements compared to those produced by baselines. The labels and con-nections between components are more accurate, suggesting that the refinement process improves the faithfulness and clarity of the diagrams. In scenarios where specific architectural details or complex data are necessary, the refined versions are closer to the original intent of diagram. This is particularly evident where GPT4-o with refinement accurately captures the necessary components and their interactions, unlike the outputs from baselines, which sometimes present a cluttered visualization.

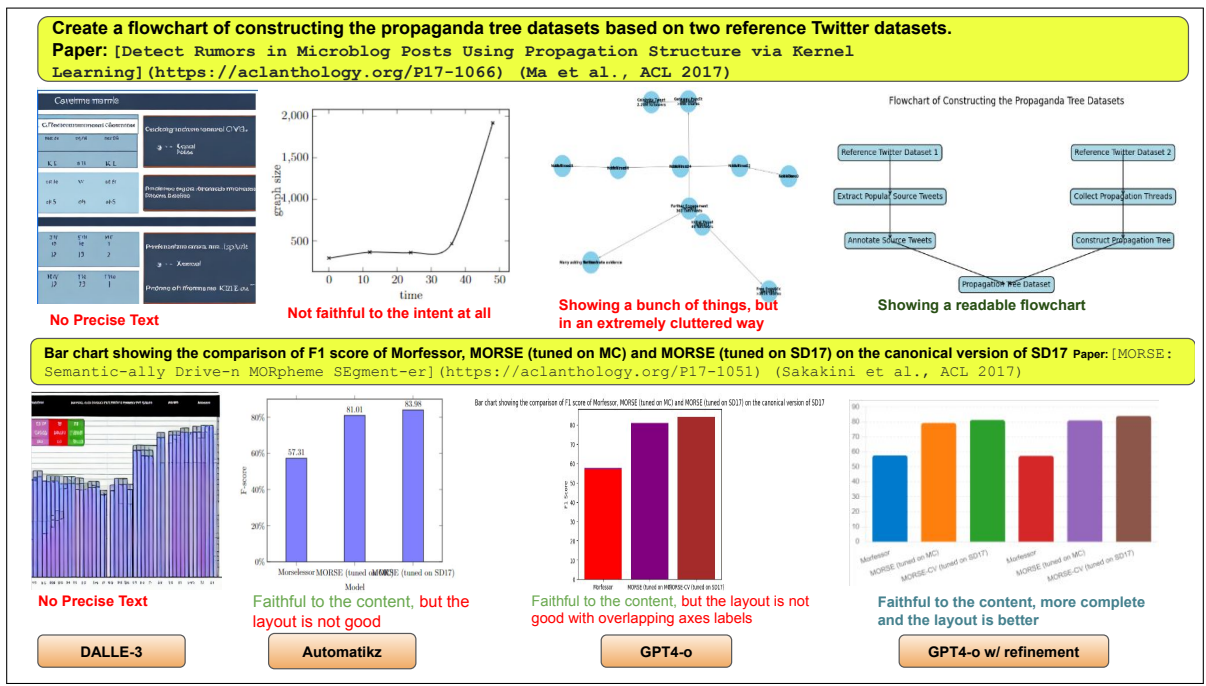

Figure 6: Qualitatively examines the images generated by different model-variants, where *we show the incompleteness and non-faithfulness errors from our baseline models and how refinement improves the quality*.

## 7 Related Work

Our work connects to several distinct but interrelated fields, so we provide a comprehensive review of the most relevant prior work.

**Text-to-Image Generation.** Text-to-image generation models have seen immense success in recent years (Ramesh et al., 2022b; Zala et al., 2023; Rombach et al., 2022b; Feng et al., 2023), but a handful of them support text conditioning (Belouadi et al., 2023) on scientific images, and to the best of our knowledge, none of these explored challenges to incorporating in a long-text scenario.

**Learning from Feedback.** Madaan et al. (2023) first introduced 'Self-Refine' and then there had been multiple works exploring the use of natural language feedback to improve performance. Lee et al. (2024a) proposes a framework to reduce hallucination in question answering. Ki and Carpuat (2024) exploits the strengths of LLMs and machine translation by guiding LLMs to automatically post-edit MT with external feedback on its quality. Lee et al. (2024b) introduced a multimodal self-feedback guided revision model to reduce hallucination. The work most closely related to ours is that of Nathani et al. (2023). However, our approach diverges from theirs in a key aspect: our feedback modules are uniquely grounded in cross-verifying information from long texts. Furthermore, we introduce a novel algorithm that refines this process sequentially, which has proven to be more effective in our specific use case. We also discuss the related works on code generation in Appendix 18.

## 8 Discussion and Future Work

We introduce an innovative task for constructing scientific diagrams from scientific document discourse, commonly used in presentations/posters. We establish a new benchmark using document-to-slides dataset and a sequential diagram refinement-based methodology to generate diagrams from academic papers. In future, we would like to assess the granularity of feedback that helps in diagram improvement by designing rubrics about what a good diagram should look like by communicating with the experts in the data visualization communities.

## Acknowledgement

We thank the anonymous ARR reviewers and UMD CLIP members for their constructive comments and feedback on the draft. This material is based upon work supported by the National Science Foundation under Grant No. IIS-2403436 (Boyd-Graber). Any opinions, findings, and conclusions or recommendations expressed in this material are those of the author(s) and do not necessarily reflect the views of the National Science Foundation. Annotations and support for Ishani Mondal were provided by a gift from Adobe Research.

## Limitations

While our SciDoc2Diagrammer-MAF represents a significant advancement in automated diagram generation, some limitations should be noted:

1. **Small Dataset Size.** The task of creating diagrams that are both factually correct and visually appealing, based on long academic papers and user intent, requires deep domain knowledge and expert input. In the construction of SciDoc2DiagramBench, expert human annotators with extensive experience in natural language processing (NLP) and diagram creation were employed to ensure that the diagrams met the required standards of completeness, faithfulness, and layout quality. Hence, the current version of SciDoc2DiagramBench includes 1080 diagrams from 89 ACL papers, which, although diverse, is relatively small.

2. **Complexity in Layout Refinement.** As noted, our model struggles with improving the layout during the refinement process. This is particularly evident in complex visualizations where spatial and design elements are crucial. The current algorithms may not adequately capture aesthetic and functional aspects of layout design, impacting the overall clarity and effectiveness of the diagrams.

3. **Feedback Mechanism Limitations.** While the multi-aspect feedback mechanism aims to enhance diagram quality by focusing on completeness, faithfulness, and layout, the effectiveness of this feedback depends on the precision of the underlying evaluative criteria. Misalignments in feedback interpretation can lead to overfitting on specific diagram aspects at the expense of others.

## Ethics Statement

The experiments performed in this study involved human participants. All the experiments involving human evaluation were exempt under institutional IRB review. We recruited participants for our human study using Upwork and we have fairly compensated all the Upwork freelancers involved in this study, at an average rate of 20.00 USD per hour (respecting their suggested Upwork hourly wage). We did not collect any personal data during the experiments, and they could choose not to participate in the study. The documents used in the study are distributed under an open license.

## References

2008–2024. Grobid. https://github.com/kermitt2/grobid.

Marah Abdin, Sam Ade Jacobs, Ammar Ahmad Awan, Jyoti Aneja, Ahmed Awadallah, Hany Awadalla, Nguyen Bach, Amit Bahree, Arash Bakhtiari, Harkirat Behl, et al. 2024. Phi-3 technical report: A highly capable language model locally on your phone. *arXiv preprint arXiv:2404.14219*.

Yoshia Abe, Tatsuya Daikoku, and Yasuo Kuniyoshi. 2024. Assessing the aesthetic evaluation capabilities of gpt-4 with vision: Insights from group and individual assessments.

Aishwarya Agrawal, Jiasen Lu, Stanislaw Antol, Margaret Mitchell, C. Lawrence Zitnick, Dhruv Batra, and Devi Parikh. 2016. Vqa: Visual question answering.

Haozhe An, Christabel Acquaye, Colin Wang, Zongxia Li, and Rachel Rudinger. 2024. Do large language models discriminate in hiring decisions on the basis of race, ethnicity, and gender?

Sören Auer, Dante AC Barone, Cassiano Bartz, Eduardo G Cortes, Mohamad Yaser Jaradeh, Oliver Karras, Manolis Koubarakis, Dmitry Mouromtsev, Dmitrii Pliukhin, Daniil Radyush, et al. 2023. The sciqa scientific question answering benchmark for scholarly knowledge. *Scientific Reports*, 13(1):7240.

Jonas Belouadi, Anne Lauscher, and Steffen Eger. 2023. Automatikz: Text-guided synthesis of scientific vector graphics with tikz. *ArXiv*, abs/2310.00367.

Mimi V Chapman, William J Hall, Robert Colby, and Laurel AG Sisler. 2014. How images work: An analysis of a visual intervention used to facilitate a difficult conversation and promote understanding. *Qualitative Social Work*, 13(4):456–476.

Ting-Rui Chiang and Yun-Nung Chen. 2019. Semantically-aligned equation generation for solving and reasoning math word problems. In

*Proceedings of the 2019 Conference of the North American Chapter of the Association for Computational Linguistics: Human Language Technologies, Volume 1 (Long and Short Papers)*, pages 2656–2668, Minneapolis, Minnesota. Association for Computational Linguistics.

Christopher Clark and Santosh Kumar Divvala. 2016. Pdffigures 2.0: Mining figures from research papers. *2016 IEEE/ACM Joint Conference on Digital Libraries (JCDL)*, pages 143–152.

Alexander Fabbri, Irene Li, Prawat Trairatvorakul, Yijiao He, Weitai Ting, Robert Tung, Caitlin Westerfield, and Dragomir Radev. 2018. TutorialBank: A manually-collected corpus for prerequisite chains, survey extraction and resource recommendation. In *Proceedings of the 56th Annual Meeting of the Association for Computational Linguistics (Volume 1: Long Papers)*, pages 611–620, Melbourne, Australia. Association for Computational Linguistics.

Weixi Feng, Wanrong Zhu, Tsu jui Fu, Varun Jampani, Arjun Akula, Xuehai He, Sugato Basu, Xin Eric Wang, and William Yang Wang. 2023. Layoutgpt: Compositional visual planning and generation with large language models.

Tsu-Jui Fu, William Yang Wang, Daniel J. McDuff, and Yale Song. 2021. Doc2ppt: Automatic presentation slides generation from scientific documents. In *AAAI Conference on Artificial Intelligence*.

Hanan Gani, Shariq Farooq Bhat, Muzammal Naseer, Salman Khan, and Peter Wonka. 2024. Llm blueprint: Enabling text-to-image generation with complex and detailed prompts.

Jack Hessel, Ari Holtzman, Maxwell Forbes, Ronan Le Bras, and Yejin Choi. 2021. CLIPScore: A reference-free evaluation metric for image captioning. In *Proceedings of the 2021 Conference on Empirical Methods in Natural Language Processing*, pages 7514–7528, Online and Punta Cana, Dominican Republic. Association for Computational Linguistics.

Ting-Yao Hsu, C Lee Giles, and Ting-Hao Huang. 2021. SciCap: Generating captions for scientific figures. In *Findings of the Association for Computational Linguistics: EMNLP 2021*, pages 3258–3264, Punta Cana, Dominican Republic. Association for Computational Linguistics.

Lei Huang, Weijiang Yu, Weitao Ma, Weihong Zhong, Zhangyin Feng, Haotian Wang, Qianglong Chen, Weihua Peng, Xiaocheng Feng, Bing Qin, and Ting Liu. 2023. A survey on hallucination in large language models: Principles, taxonomy, challenges, and open questions.

Wei Huang, Xudong Ma, Haotong Qin, Xingyu Zheng, Chengtao Lv, Hong Chen, Jie Luo, Xiaojuan Qi, Xianglong Liu, and Michele Magno. 2024. How good are low-bit quantized llama3 models? an empirical study.

Albert Q. Jiang, Alexandre Sablayrolles, Arthur Mensch, Chris Bamford, Devendra Singh Chaplot, Diego de las Casas, Florian Bressand, Gianna Lengyel, Guillaume Lample, Lucile Saulnier, Lélio Renard Lavaud, Marie-Anne Lachaux, Pierre Stock, Teven Le Scao, Thibaut Lavril, Thomas Wang, Timothée Lacroix, and William El Sayed. 2023. Mistral 7b.

Shankar Kantharaj, Rixie Tiffany Leong, Xiang Lin, Ahmed Masry, Megh Thakkar, Enamul Hoque, and Shafiq Joty. 2022. Chart-to-text: A large-scale benchmark for chart summarization. In *Proceedings of the 60th Annual Meeting of the Association for Computational Linguistics (Volume 1: Long Papers)*, pages 4005–4023, Dublin, Ireland. Association for Computational Linguistics.

Dayeon Ki and Marine Carpuat. 2024. Guiding large language models to post-edit machine translation with error annotations.

Anastasia Krithara, Anastasios Nentidis, Konstantinos Bougiatiotis, and Georgios Paliouras. 2023. Bioasq-qa: A manually curated corpus for biomedical question answering. *Scientific Data*, 10(1):170.

Dongyub Lee, Eunhwan Park, Hodong Lee, and Heuiseok Lim. 2024a. Ask, assess, and refine: Rectifying factual consistency and hallucination in LLMs with metric-guided feedback learning. In *Proceedings of the 18th Conference of the European Chapter of the Association for Computational Linguistics (Volume 1: Long Papers)*, pages 2422–2433, St. Julian's, Malta. Association for Computational Linguistics.

Seongyun Lee, Hyunjae Kim, and Jaewoo Kang. 2023. Liquid: a framework for list question answering dataset generation. In *Proceedings of the AAAI Conference on Artificial Intelligence*, volume 37, pages 13014–13024.

Seongyun Lee, Sue Hyun Park, Yongrae Jo, and Minjoon Seo. 2024b. Volcano: Mitigating multimodal hallucination through self-feedback guided revision.

Patrick Lewis, Ethan Perez, Aleksandara Piktus, Fabio Petroni, Vladimir Karpukhin, Naman Goyal, Heinrich Kuttler, Mike Lewis, Wen tau Yih, Tim Rocktäschel, Sebastian Riedel, and Douwe Kiela. 2020. Retrieval-augmented generation for knowledge-intensive nlp tasks. *ArXiv*, abs/2005.11401.

Kaixin Li, Yuchen Tian, Qisheng Hu, Ziyang Luo, and Jing Ma. 2024a. Mmcode: Evaluating multi-modal code large language models with visually rich programming problems.

Zongxia Li, Andrew Mao, Daniel Stephens, Pranav Goel, Emily Walpole, Alden Dima, Juan Fung, and Jordan Boyd-Graber. 2024b. Improving the TENOR of labeling: Re-evaluating topic models for content analysis. In *Proceedings of the 18th Conference of the European Chapter of the Association for Computational Linguistics (Volume 1: Long Papers)*, pages 840–859, St. Julian's, Malta. Association for Computational Linguistics.

Zongxia Li, Ishani Mondal, Yijun Liang, Huy Nghiem, and Jordan Lee Boyd-Graber. 2024c. Pedants: Cheap but effective and interpretable answer equivalence.

Chin-Yew Lin. 2004. ROUGE: A package for automatic evaluation of summaries. In *Text Summarization Branches Out*, pages 74–81, Barcelona, Spain. Association for Computational Linguistics.

Fangyu Liu, Julian Eisenschlos, Francesco Piccinno, Syrine Krichene, Chenxi Pang, Kenton Lee, Mandar Joshi, Wenhu Chen, Nigel Collier, and Yasemin Altun. 2023a. DePlot: One-shot visual language reasoning by plot-to-table translation. In *Findings of the Association for Computational Linguistics: ACL 2023*, pages 10381–10399, Toronto, Canada. Association for Computational Linguistics.

Fangyu Liu, Francesco Piccinno, Syrine Krichene, Chenxi Pang, Kenton Lee, Mandar Joshi, Yasemin Altun, Nigel Collier, and Julian Eisenschlos. 2023b. MatCha: Enhancing visual language pretraining with math reasoning and chart derendering. In *Proceedings of the 61st Annual Meeting of the Association for Computational Linguistics (Volume 1: Long Papers)*, pages 12756–12770, Toronto, Canada. Association for Computational Linguistics.

Jing Ma, Wei Gao, and Kam-Fai Wong. 2017. Detect rumors in microblog posts using propagation structure via kernel learning. In *Proceedings of the 55th Annual Meeting of the Association for Computational Linguistics (Volume 1: Long Papers)*, pages 708–717, Vancouver, Canada. Association for Computational Linguistics.

Aman Madaan, Niket Tandon, Prakhar Gupta, Skyler Hallinan, Luyu Gao, Sarah Wiegreffe, Uri Alon, Nouha Dziri, Shrimai Prabhumoye, Yiming Yang, Shashank Gupta, Bodhisattwa Prasad Majumder, Katherine Hermann, Sean Welleck, Amir Yazdanbakhsh, and Peter Clark. 2023. Self-refine: Iterative refinement with self-feedback.

Paula Maddigan and Teo Susnjak. 2023. Chat2vis: Generating data visualisations via natural language using chatgpt, codex and gpt-3 large language models. *Ieee Access*.

Ahmed Masry, Parsa Kavehzadeh, Xuan Long Do, Enamul Hoque, and Shafiq Joty. 2023. UniChart: A universal vision-language pretrained model for chart comprehension and reasoning. In *Proceedings of the 2023 Conference on Empirical Methods in Natural Language Processing*, pages 14662–14684, Singapore. Association for Computational Linguistics.

Ishani Mondal, Yufang Hou, and Charles Jochim. 2021. End-to-end construction of NLP knowledge graph. In *Findings of the Association for Computational Linguistics: ACL-IJCNLP 2021*, pages 1885–1895, Online. Association for Computational Linguistics.

Ishani Mondal, Shwetha S, Anandhavelu Natarajan, Aparna Garimella, Sambaran Bandyopadhyay, and Jordan Boyd-Graber. 2024. Presentations by the humans and for the humans: Harnessing LLMs for generating persona-aware slides from documents. In *Proceedings of the 18th Conference of the European Chapter of the Association for Computational Linguistics (Volume 1: Long Papers)*, pages 2664–2684, St. Julian's, Malta. Association for Computational Linguistics.

Kazuma Murao, Ken Kobayashi, Hayato Kobayashi, Taichi Yatsuka, Takeshi Masuyama, Tatsuru Higurashi, and Yoshimune Tabuchi. 2019. A case study on neural headline generation for editing support. In *Proceedings of the 2019 Conference of the North American Chapter of the Association for Computational Linguistics: Human Language Technologies, Volume 2 (Industry Papers)*, pages 73–82, Minneapolis, Minnesota. Association for Computational Linguistics.

Deepak Nathani, David Wang, Liangming Pan, and William Wang. 2023. MAF: Multi-aspect feedback for improving reasoning in large language models. In *Proceedings of the 2023 Conference on Empirical Methods in Natural Language Processing*, pages 6591–6616, Singapore. Association for Computational Linguistics.

Huy Nghiem, John Prindle, Jieyu Zhao, and Hal Daumé III au2. 2024. "you gotta be a doctor, lin": An investigation of name-based bias of large language models in employment recommendations.

R OpenAI. 2023. Gpt-4 technical report. *ArXiv*, 2303.

Lin Qiu, Hao Zhou, Yanru Qu, Weinan Zhang, Suoheng Li, Shu Rong, Dongyu Ru, Lihua Qian, Kewei Tu, and Yong Yu. 2018. Qa4ie: A question answering based framework for information extraction. In *The Semantic Web–ISWC 2018: 17th International Semantic Web Conference, Monterey, CA, USA, October 8–12, 2018, Proceedings, Part I 17*, pages 198–216. Springer.

Aditya Ramesh, Prafulla Dhariwal, Alex Nichol, Casey Chu, and Mark Chen. 2022a. Hierarchical text-conditional image generation with clip latents. *ArXiv*, abs/2204.06125.

Aditya Ramesh, Prafulla Dhariwal, Alex Nichol, Casey Chu, and Mark Chen. 2022b. Hierarchical text-conditional image generation with clip latents.

Aditya Ramesh, Mikhail Pavlov, Gabriel Goh, Scott Gray, Chelsea Voss, Alec Radford, Mark Chen, and Ilya Sutskever. 2021. Zero-shot text-to-image generation. In *Proceedings of the 38th International Conference on Machine Learning*, volume 139 of *Proceedings of Machine Learning Research*, pages 8821–8831. PMLR.

Marco Tulio Ribeiro, Tongshuang Wu, Carlos Guestrin, and Sameer Singh. 2020. Beyond accuracy: Behavioral testing of NLP models with CheckList. In *Proceedings of the 58th Annual Meeting of the Association for Computational Linguistics*, pages 4902–4912, Online. Association for Computational Linguistics.

R. Rombach, A. Blattmann, D. Lorenz, P. Esser, and B. Ommer. 2022a. High-resolution image synthesis with latent diffusion models. In *2022 IEEE/CVF Conference on Computer Vision and Pattern Recognition (CVPR)*, pages 10674–10685, Los Alamitos, CA, USA. IEEE Computer Society.

Robin Rombach, A. Blattmann, Dominik Lorenz, Patrick Esser, and Björn Ommer. 2021. High-resolution image synthesis with latent diffusion models. *2022 IEEE/CVF Conference on Computer Vision and Pattern Recognition (CVPR)*, pages 10674–10685.

Robin Rombach, Andreas Blattmann, Dominik Lorenz, Patrick Esser, and Björn Ommer. 2022b. High-resolution image synthesis with latent diffusion models. In *Proceedings of the IEEE/CVF Conference on Computer Vision and Pattern Recognition (CVPR)*, pages 10684–10695.

Baptiste Rozière, Jonas Gehring, Fabian Gloeckle, Sten Sootla, Itai Gat, Xiaoqing Ellen Tan, Yossi Adi, Jingyu Liu, Romain Sauvestre, Tal Remez, Jérémy Rapin, Artyom Kozhevnikov, Ivan Evtimov, Joanna Bitton, Manish Bhatt, Cristian Canton Ferrer, Aaron Grattafiori, Wenhan Xiong, Alexandre Défossez, Jade Copet, Faisal Azhar, Hugo Touvron, Louis Martin, Nicolas Usunier, Thomas Scialom, and Gabriel Synnaeve. 2024. Code llama: Open foundation models for code.

Tarek Sakakini, Suma Bhat, and Pramod Viswanath. 2017. MORSE: Semantic-ally drive-n MORpheme SEgment-er. In *Proceedings of the 55th Annual Meeting of the Association for Computational Linguistics (Volume 1: Long Papers)*, pages 552–561, Vancouver, Canada. Association for Computational Linguistics.

Shreya Shankar, J. D. Zamfirescu-Pereira, Björn Hartmann, Aditya G. Parameswaran, and Ian Arawjo. 2024. Who validates the validators? aligning llm-assisted evaluation of llm outputs with human preferences.

Chufan Shi, Cheng Yang, Yaxin Liu, Bo Shui, Junjie Wang, Mohan Jing, Linran Xu, Xinyu Zhu, Siheng Li, Yuxiang Zhang, Gongye Liu, Xiaomei Nie, Deng Cai, and Yujiu Yang. 2024. Chartmimic: Evaluating lmm's cross-modal reasoning capability via chart-to-code generation.

Chenglei Si, Yanzhe Zhang, Zhengyuan Yang, Ruibo Liu, and Diyi Yang. 2024. Design2code: How far are we from automating front-end engineering?

Edward Sun, Yufang Hou, Dakuo Wang, Yunfeng Zhang, and Nancy X. R. Wang. 2021. D2S: Document-to-slide generation via query-based text summarization. In *Proceedings of the 2021 Conference of the North American Chapter of the Association for Computational Linguistics: Human Language Technologies*, pages 1405–1418, Online. Association for Computational Linguistics.

Yi Tay, Anh Tuan Luu, Siu Cheung Hui, and Jian Su. 2018. Reasoning with sarcasm by reading in-between. In *Proceedings of the 56th Annual Meeting of the Association for Computational Linguistics (Volume 1: Long Papers)*, pages 1010–1020, Melbourne, Australia. Association for Computational Linguistics.

Hugo Touvron, Louis Martin, Kevin Stone, Peter Albert, Amjad Almahairi, Yasmine Babaei, Nikolay Bashlykov, Soumya Batra, Prajjwal Bhargava, Shruti Bhosale, Dan Bikel, Lukas Blecher, Cristian Canton Ferrer, Moya Chen, Guillem Cucurull, David Esiobu, Jude Fernandes, Jeremy Fu, Wenyin Fu, Brian Fuller, Cynthia Gao, Vedanuj Goswami, Naman Goyal, Anthony Hartshorn, Saghar Hosseini, Rui Hou, Hakan Inan, Marcin Kardas, Viktor Kerkez, Madian Khabsa, Isabel Kloumann, Artem Korenev, Punit Singh Koura, Marie-Anne Lachaux, Thibaut Lavril, Jenya Lee, Diana Liskovich, Yinghai Lu, Yuning Mao, Xavier Martinet, Todor Mihaylov, Pushkar Mishra, Igor Molybog, Yixin Nie, Andrew Poulton, Jeremy Reizenstein, Rashi Rungta, Kalyan Saladi, Alan Schelten, Ruan Silva, Eric Michael Smith, Ranjan Subramanian, Xiaoqing Ellen Tan, Binh Tang, Ross Taylor, Adina Williams, Jian Xiang Kuan, Puxin Xu, Zheng Yan, Iliyan Zarov, Yuchen Zhang, Angela Fan, Melanie Kambadur, Sharan Narang, Aurelien Rodriguez, Robert Stojnic, Sergey Edunov, and Thomas Scialom. 2023. Llama 2: Open foundation and fine-tuned chat models.

Shubham Ugare, Tarun Suresh, Hangoo Kang, Sasa Misailovic, and Gagandeep Singh. 2024. Improving llm code generation with grammar augmentation. *arXiv preprint arXiv:2403.01632*.

Yifan Yao, Jinhao Duan, Kaidi Xu, Yuanfang Cai, Zhibo Sun, and Yue Zhang. 2024. A survey on large language model (llm) security and privacy: The good, the bad, and the ugly. *High-Confidence Computing*, 4(2):100211.

Abhay Zala, Han Lin, Jaemin Cho, and Mohit Bansal. 2023. Diagrammergpt: Generating open-domain, open-platform diagrams via llm planning.

Tianyi Zhang, Varsha Kishore, Felix Wu, Kilian Q. Weinberger, and Yoav Artzi. 2019. Bertscore: Evaluating text generation with BERT. *CoRR*, abs/1904.09675.

Xinlu Zhang, Yujie Lu, Weizhi Wang, An Yan, Jun Yan, Lianke Qin, Heng Wang, Xifeng Yan, William Yang Wang, and Linda Ruth Petzold. 2023. Gpt-4v(ision) as a generalist evaluator for vision-language tasks.

Tiancheng Zhao, Ran Zhao, and Maxine Eskenazi. 2017. Learning discourse-level diversity for neural dialog models using conditional variational autoencoders.

In *Proceedings of the 55th Annual Meeting of the Association for Computational Linguistics (Volume 1: Long Papers)*, pages 654–664, Vancouver, Canada. Association for Computational Linguistics.

Yuhang Zhou, Jing Zhu, Paiheng Xu, Xiaoyu Liu, Xiyao Wang, Danai Koutra, Wei Ai, and Furong Huang. 2024. Multi-stage balanced distillation: Addressing long-tail challenges in sequence-level knowledge distillation. *arXiv preprint arXiv:2406.13114*.

## 9 User-Study Guided Motivation of Task

**Why do we need this Task?** Our motivation is primarily inspired by the SlideGeneration dataset (Fu et al., 2021), which we have already cited in Footnote 1 of the paper, indicating user's need. However, to better motivate our task, we survey 11 participants (Instructions in Figure 7), ranging from beginners to experts in NLP, comprising of 6 MS/PhD students with at least 1 *CL or ML conference paper, 2 people from ML research lab, 3 beginners in NLP. We asked them to review 3 *CL papers each and determine if additional visuals were needed for slide presentations *"Would you like to create some more figures that you think would be required for better visualization in the slides, which are not present inside the paper directly? If yes, can you provide some description of these additional images?"* Out of 33 responses (11*3) from the papers, 28 answers (84.8%) indicate the need for additional visualization in slides.

The survey shows that each paper requires an average of 2.3 "Extrapolated-Diagrams", with a focus on flowcharts, converting complex tables into charts, and synthesizing related work to highlight novelty. These findings indicate the potential of including such diagrams in our new dataset, which aligns with the additional visuals found in (Fu et al., 2021).

## 10 Effectiveness of our DiagramGeneration Pipeline

Based on the survey responses in 9, we selected the paper requiring the most number of "Extrapolated-Diagrams" and asked two participants from the set of 11 people (Control Group) to create a flowchart and a bar chart from scratch after reading the paper content. On average, it took them 30-35 minutes per diagram. We then provided two other participants (Experimental Group) with access to our interactive SciDoc2Diagrammer-MAF system to create the same set of diagrams (a flowchart and a bar chart). With our tool, they completed each diagram in 8-10 minutes. We also provided intermediate questions and answers to extract data from the paper (Figure 8), code (Figure 9) which helped reduce the cognitive load by simplifying the process of diagram creation. The participants found it easier to tweak a template diagram generated through code, instead of searching for the relevant data from paper, and writing a code from scratch to generate figures (Sample shown in (Figure 10)).

## 11 Hiring Upwork Participants

### 11.1 Hiring Workers for Dataset Creation

Using Upwork, we hired three workers familiar with Machine learning and NLP with almost 5 years of experience and well-versed with creating presentations from documents, sorted by having a skill set of Presentation making. The hiring was made after shortlisting them through interviews, where they were initially asked to read the paper (**?**) and answer questions like : 1) What is the novelty of this approach? 2) What is the motivation behind the main algorithm? 3) What are the strengths and weaknesses of this paper? 4) What was the state-of-art algorithm before this model came in? 5) What kind of evaluation has been made using this approach? Moreover, they were asked to make a presentation suitable for presenting it in an AI conference. Based on their answers and the quality of the presentation being made, the first two authors of the paper made a hiring decision.

## 12 Instruction Guidelines on Annotating and Identifying Extrapolated Diagrams

**Objective:** You are provided with a dataset containing scientific papers and their corresponding slide decks. Your task is to annotate each slide in the paper according to predefined categories based on the content origin and presentation style.

### 12.1 Categories:

1. **Mere Extraction from the Paper**: Slides that directly copy and paste content from the paper.
2. **Extrapolated Diagrams**: Slides that include more elaborative visualizations or explanations beyond mere copying. This includes:
    - Diagrams with textual explanations, highlights, or key takeaways.

Figure 7: User Study Survey Form to Understand how many people would like to create new Figures from a few selected *CL papers while creating presentations

- Flowcharts that summarize or clarify the content.
- Examples that illustrate theoretical concepts.
- Conversions of tabular data into graphical representations.

3. **External**: Slides containing content sourced from external references outside of the paper.

**Procedure:**

1. **Initial Check**: Begin by reviewing the slide deck to determine if there are any "Extrapolated Diagrams". Only proceed with the annotation if such slides are present.
2. **Comprehensive Review**: Read through the entire paper to understand the context and origin of images and diagrams used in the slides.
3. **Annotation**: Classify each slide into one of the categories based on your understanding and findings. Consider the source and nature of the content on each slide.

Ensure accuracy in your annotations by thoroughly comparing slide content with the original paper and other sources mentioned within the slides.

## 13 Instructions on Annotating Intent of the Diagram Creation

**Objective:** Your role is to provide clear, comprehensive, and complete descriptions for diagrams in scientific slide decks, focusing on their intent and content. This will enable any individual to understand or recreate the diagram based purely on your description.

**Steps to Follow:**

1. **Identify the Diagram:** Start by identifying whether the slide includes bar charts, plots, or tables. Note the context in which these diagrams are used, such as summarizing results or discussing related works.
2. **Determine the Intent:**
   - What is the diagram illustrating? Understand what the main point or data the diagram conveys.
   - Why is this diagram used? Analyze why the author chose this type of diagram to present the data.
   - How does the diagram achieve its purpose? Look at the elements of the diagram like axes, labels, legends, and how they contribute to conveying the intended message.

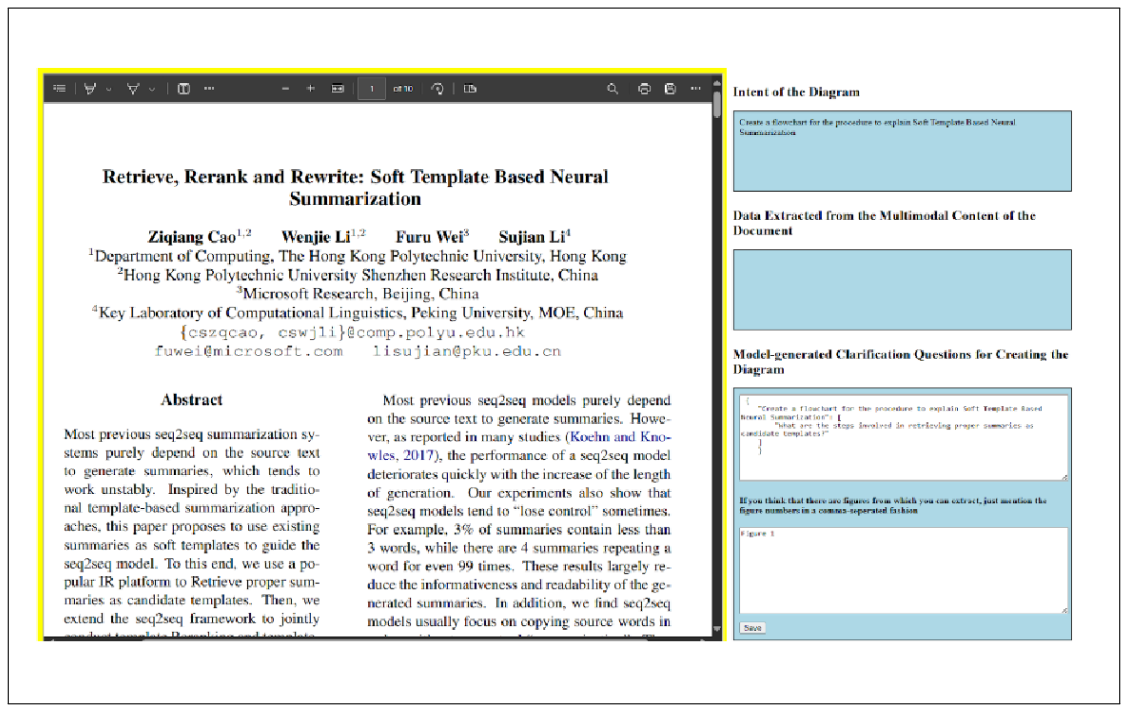

Figure 8: Shows the Page 1 of the DiagramGeneration Interface. The left panel displays a scientific document titled "Retrieve, Rerank and Rewrite: Soft Template Based Neural Summarization," including the title, authors, affiliations, and abstract. The right panel shows the interface used to generate diagrams based on user intent. The first section captures the "Intent of the Diagram," which in this case is "Create a flowchart for the procedure to explain Soft Template Based Neural Summarization." The second section includes "Data Extracted from the Multimodal Content of the Document," while the third section displays "Model-generated Clarification Questions for Creating the Diagram," asking for additional information such as figure names in comma-separated format. This layout illustrates a pipeline where user input and model clarifications are integrated to guide the generation of scientific diagrams.

3. **Write the Description:**
   - Your description should allow someone who hasn't seen the diagram to visualize and possibly recreate it accurately.
   - Include details about:
     - **Data Representation:** What data is being represented and how?
     - **Layout and Design:** Describe the layout, color scheme, and any textual elements.
     - **Interpretative Elements:** Explain any elements that help in interpreting the data, such as legends, scales, or reference lines.
4. **Verify Completeness:** Review your description to ensure it captures all aspects of the diagram comprehensively.

**Example of a Good Intent:**

"Create a bar chart titled 'Annual Growth' displaying the yearly sales growth from 2015 to 2020. Each bar represents a year, colored in gradient from dark blue (2015) to light blue (2020). The Y-axis is labeled 'Percentage Growth' and ranges from 0% to 50%, increasing by 5% increments. A legend in the bottom right explains the color gradient as representing successive years."

**Note:** Ensure that your descriptions are free of assumptions and interpretations beyond what is explicitly shown in the diagrams.

## 14 Addressing Data Leakage Concerns

We address the potential data leakage due to the use of existing datasets such as DOC2PPT (Fu et al., 2021) and TutorialBank (Fabbri et al., 2018) in constructing the SciDoc2DiagramBench by clarifying that our task is fundamentally distinct from these datasets' original objectives, such as slide creation. Specifically, SciDoc2Diagram aims to evaluate large language models' (LLMs) ability to generate scientific diagrams conditioned on user intent and long-document context, which presents a more nuanced and challenging task than the straightforward generation of slides from text.

Our task focuses on creating scientific diagrams based on detailed user instructions and content from the academic paper itself. This involves synthesizing complex ideas into visual form, which is substantially different from simply converting scientific text into slides. As such, the focus shifts from summarization or slide creation to a more advanced task of diagrammatic representation, ensuring that any overlap with DOC2PPT (Fu et al., 2021) or TutorialBank (Fabbri et al., 2018) is minimized. Additionally, no direct train-test splits are used, and our dataset is crafted to assess the ability to generalize across varied diagrammatic intents.

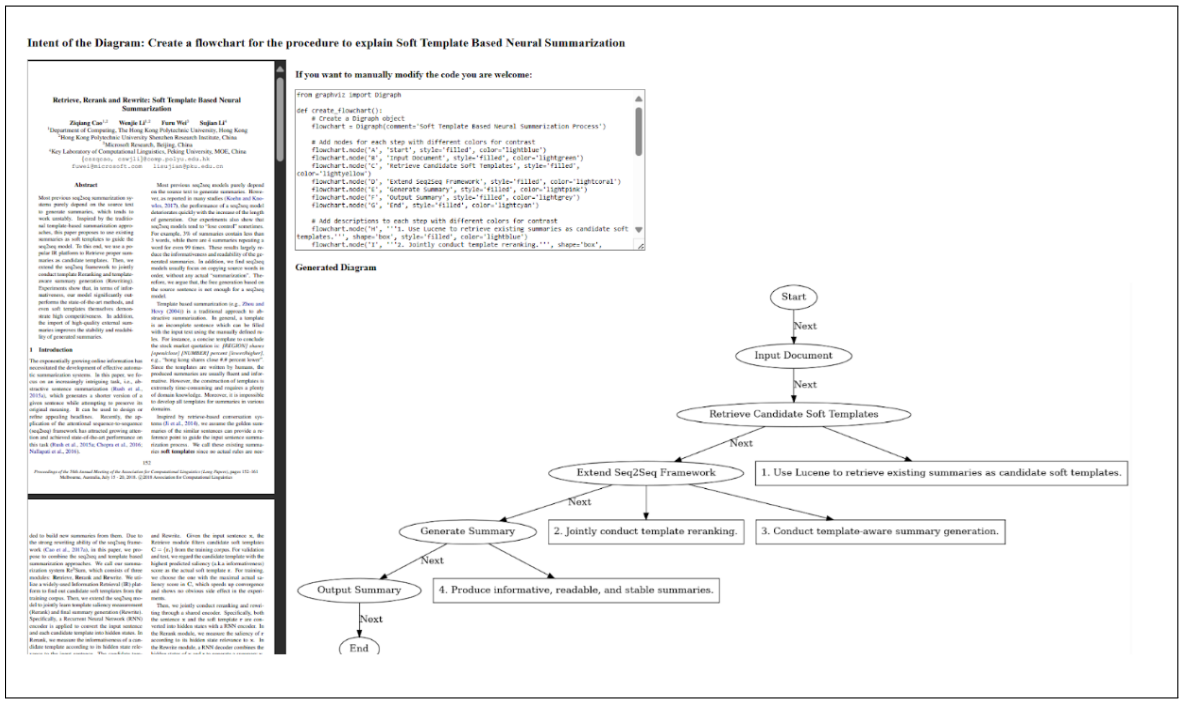

Figure 9: shows the Page 2 of DiagramGeneration Interface. The left panel shows a scientific paper titled "Retrieve, Rerank and Rewrite: Soft Template Based Neural Summarization," including the abstract and introduction sections. The middle panel contains a code snippet that represents a Graphviz script for creating the diagram. The script allows users to manually modify the code to customize the flowchart. The right panel displays the generated flowchart based on the provided code. The flowchart visualizes the steps involved in the summarization process.

Besides, it is worth noting that many early benchmarks in the field have been developed by reusing existing datasets or open-source repositories to create new tasks. Examples include works like Automatikz (Belouadi et al., 2023), PersonaD2S (Mondal et al., 2024), and E2EKG (Mondal et al., 2021) which repurpose existing data to establish more specific, challenging benchmarks. Our work follows a similar approach, using existing resources to create a more complex benchmark aimed at evaluating how diagrams can be generated not only based on the academic paper but also in line with the specific user intent, a dimension absent in DOC2PPT and TutorialBank.

To further mitigate concerns about data leakage and to enhance the diversity and representativeness of the SciDoc2DiagramBench, we have incorporated expert human annotations. These annotations include additional metadata such as the intent behind the created diagram. The inclusion of this metadata makes each instance more unique and specifically tailored to the task of diagram creation, thereby distinguishing it from simple text-to-slide generation datasets. This human-curated layer, not present in DOC2PPT or TutorialBank, ensures that our dataset adds substantial value and differentiation from prior resources.

In conclusion, while we leverage existing datasets for constructing SciDoc2DiagramBench, the way we have formulated the task introduces a significant layer of complexity and novelty. By focusing on user intent, incorporating human annotations, and avoiding standard train-test splits, we ensure that SciDoc2DiagramBench remains a challenging and representative benchmark that builds on existing resources while addressing concerns about potential data leakage. Besides, a significant number of early researchers have created benchmarks by reusing existing datasets or open-source repositories such as Automatikz (Belouadi et al.,

[Figure showing a feedback form interface with fields for Completeness (1-5), additional questions, Correctness (1-5), correct data input area, and Regenerate/Save buttons]

Figure 10: displays a feedback form designed to allow users to evaluate and improve the generated diagram and its underlying code. Users can provide ratings for the completeness and correctness of the diagram on a scale from 1 to 5. Additionally, users can ask clarification questions to verify the completeness of the generated content by submitting their queries in a comma-separated format. They can also input correct data or corrections in the text box provided to regenerate the code. The interface includes buttons to either "Regenerate" or "Save" the revised code based on user input. This tool enables interactive refinement of generated diagrams based on user feedback.

2023), PersonaD2S (Mondal et al., 2024), End-to-End Scientific Knowledge Graph Construction (Mondal et al., 2021), VQA (Agrawal et al., 2016).

## 15 Overview of SciDoc2DiagramBench

### 15.1 Examples from SciMultiDoc2DiagramBench-Gold

Figure 13 corresponds to Intent "Construct a detailed and organized table to summarize various human-generated datasets used in natural language processing research which includes the name of each dataset, the annotators involved, the format of the data, and the size of the dataset."

Figure 14 (Extrapolated-Summary) corresponds to Intent "Construct a summary table showing the comparison of exisiting studies where you should specify the source paper, tables, KnowledgeBase and the target task on which it is based on' and the source Papers are As mentioned in the slides."

### 15.2 Examples from SciDoc2DiagramBench-Gold

Figure 15 (Extrapolated-Flowchart) corresponds to Intent "Construct a flowchart or diagram showing where the same context is provided to both humans and models, the human creates some response and the model creates other response along with the equations of precision and recall for appropriateness and diversity" and the source paper is (Zhao et al., 2017).

Figure 16 (Extrapolated-Results) corresponds to Intent "Create a Bar chart showing the comparison of F1 score of MORSE, Morfessor S+W, Morfessor S+W+L and MorphoChain against published state-of-the-art results" and the source paper is (Sakakini et al., 2017).

## 16 Details of the Evaluation Metrics used

BERTScore (Zhang et al., 2019) measures the semantic similarity between the captions generated by GPT-4V for gold-standard diagrams and those for our diagrams, assessing how well our visual representations capture the intended mean-ings. **ROUGE-1** (Lin, 2004) examines 1-gram overlap between these sets of captions, providing insight into the textual alignment with expected content. **CLIPScore** (Hessel et al., 2021) evaluates alignment between gold-standard images and corresponding GPT-4V captions for our generated images, indicating how closely our outputs match visual and contextual expectations.

| Types | Intent | Source Paper |
|---|---|---|
| Extrapolated-Flowchart | Create a flowchart to explain the process of how the proposed model understands the semantics of problems and decides which symbol to generate next." | (Chiang and Chen, 2019) |
| Extrapolated-Summary | Create a table to summarize related works in sarcasm detection, highlighting the novelty of the proposed model in terms of its ability to model contrast and incongruity | (Tay et al., 2018) |
| Extrapolated-Results | Convert the table describing the different translations of titles into a bar chart, grouping each translation by its source (Headline, Lead, Editor, OpenNMT, HybridFusion), and using color coding to highlight differences in the length and tone of each translation | (Murao et al., 2019) |

Table 5: Some Examples from SciDoc2DiagramBench-Extended

| Types | Description |
|---|---|
| Mere Extraction from the Paper | Directly copied and pasted into slides |
| Extrapolated Diagrams | 1) Explain a figure with text/highlights/takeaways, 2) Flowcharts, 3) Explanation through example, 4) Table to Graph or some other forms of visualization |
| External | Included from other sources (papers / web) |

Table 6: Types of Diagrams and Their Descriptions

Figure 11: shows a user interface where the generated flowchart is displayed at the top. The flowchart outlines the process for soft template-based neural summarization, detailing steps such as document input, retrieving candidate soft templates, extending the Seq2Seq framework, and generating a readable and stable summary. Below the diagram, the user is provided with a feedback section where they can rate the layout on a scale from 1 to 5 and provide remarks regarding the visual layout of the flowchart. In this example, the user suggests using a different contrast of colors and improving the figure's lock. At the bottom, the interface displays the code used to generate the diagram, written in Python using the Graphviz package, allowing for further edits or regeneration of the diagram based on user input. This setup encourages interactive refinement and iterative improvements.

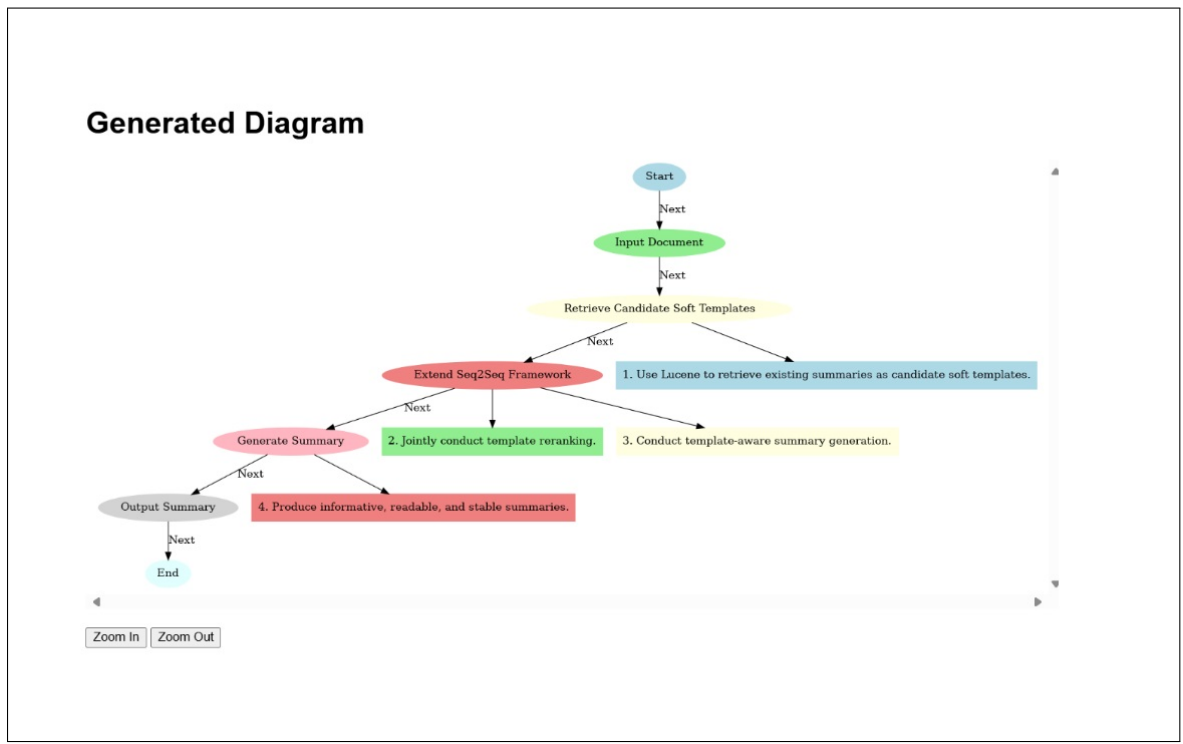

Figure 12: Page 5 of the DiagramGeneration Interface

|  | Extrapolated-Flowchart | | | Extrapolated-Results | | | Extrapolated-Architecture | | | Extrapolated-Summary | | |
| --- | --- | --- | --- | --- | --- | --- | --- | --- | --- | --- | --- | --- |
| Models (IV) | R | BS | CS | R | BS | CS | R | BS | CS | R | BS | CS |
| DALLE-3 | 0.10 | 0.23 | 0.11 | 0.03 | 0.21 | 0.23 | 0.11 | 0.25 | 0.16 | 0.03 | 0.12 | 0.13 |
| Automatikz | 0.21 | 0.28 | 0.21 | 0.12 | 0.43 | 0.17 | 0.13 | 0.54 | 0.45 | 0.49 | 0.34 | 0.27 |
| SciDoc2Diagrammer | | | | | | | | | | | | |
| w/ GPT4-o (ZS) | 0.22 | 0.36 | 0.40 | 0.23 | 0.45 | 0.29 | 0.21 | 0.46 | 0.29 | 0.28 | 0.47 | 0.50 |
| w/ GPT4-o (FS) | 0.28 | 0.47 | 0.43 | 0.33 | 0.50 | 0.38 | 0.24 | 0.48 | 0.39 | 0.42 | 0.54 | 0.52 |
| w/ GPT4-o-SR | 0.34 | 0.62 | 0.42 | 0.42 | 0.51 | 0.35 | 0.33 | 0.61 | 0.50 | 0.47 | 0.62 | 0.60 |
| w/ GPT4-o-SumMAF | 0.35 | 0.74 | 0.48 | 0.50 | 0.57 | 0.39 | 0.34 | 0.64 | 0.49 | 0.50 | 0.74 | 0.74 |
| w/ GPT4-o-SeqMAF | 0.39 | 0.79 | 0.53 | 0.49 | 0.49 | 0.36 | 0.37 | 0.58 | 0.49 | 0.45 | 0.67 | 0.62 |

Table 7: Automatic evaluation of models on various diagrams on **SciMultiDoc2DiagramBench-Gold** using ROUGE (R), BERTScore (BS), and CLIPScore (CS). The table highlights the comparative performance of DALLE-3, Automatikz, and several versions of SciDoc2Diagrammer (Few-shot Strategy) across all categories of diagrams. SciDoc2Diagrammer with GPT4-o consistently shows the best performance across most categories, indicated by the green cells, while DALLE-3 generally underperforms, as shown by the red cells.

## 17 Additional Experiments

**Individual Components are necessary in Sci-Doc2Diagrammer!** Table 9 presents the performance impact of different components in the Sci-Doc2Diagrammer model using several metrics: CLIPScore (CS), BERTScore (BS), Completeness (C), Faithfulness (F), and Layout (L). Removing the Question Generation/Answering (QG/QA) component results in significant performance drops across all metrics, highlighting its importance. The absence of the Summarization component also negatively affects the scores but to a lesser extent.

## 18 Comparison with Contemporary Research on code generation

Unlike ChartMimic (Shi et al., 2024) and MM-Code (Li et al., 2024a), which primarily focus on either visual charts or programming challenges as input, our approach uses the high-level intent pro-vided by the user to generate complex scientific diagrams. This makes our work distinct as it emphasizes understanding longer document contexts and user-specific requests rather than predefined visual or mathematical data.

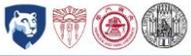

Figure 13: An Example from SciMultiDoc2DiagramBench-Gold where the Intent is "Construct a detailed and organized table to summarize various human-generated datasets used in natural language processing research which includes the name of each dataset, the annotators involved, the format of the data, and the size of the dataset." and the source Papers are As mentioned in the slides.

Figure 14: An Example from SciMultiDoc2DiagramBench-Gold where the Intent is "Construct a summary table showing the comparison of exisiting studies where you should specify the source paper, tables, KnowledgeBase and the target task on which it is based on' and the source Papers are As mentioned in the slides."

Existing works such as MMCode (Li et al., 2024a) and Design2Code (Si et al., 2024) operate on well-defined inputs like programming challenges or website designs. In contrast, our approach supports the processing of long scientific

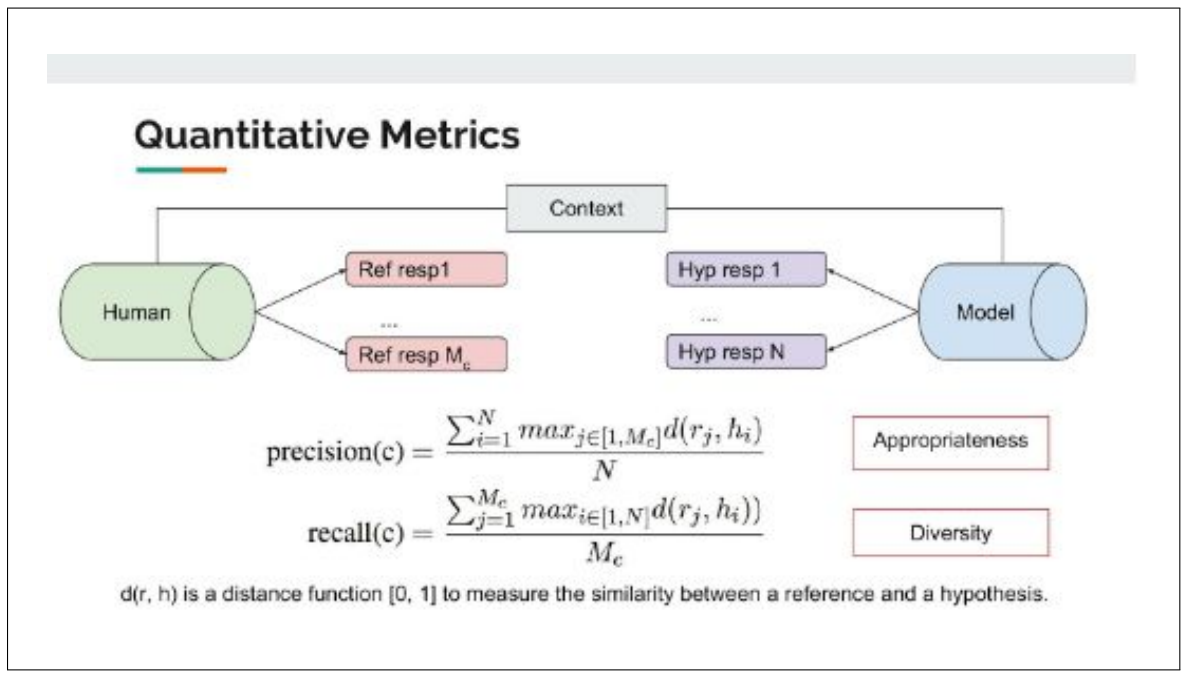

Figure 15: An Example from SciDoc2DiagramBench-Gold where the Intent is "Construct a flowchart or diagram showing where the same context is provided to both humans and models, the human creates some response and the model creates other response along with the equations of precision and recall for appropriateness and diversity .." and the source paper is (Zhao et al., 2017)

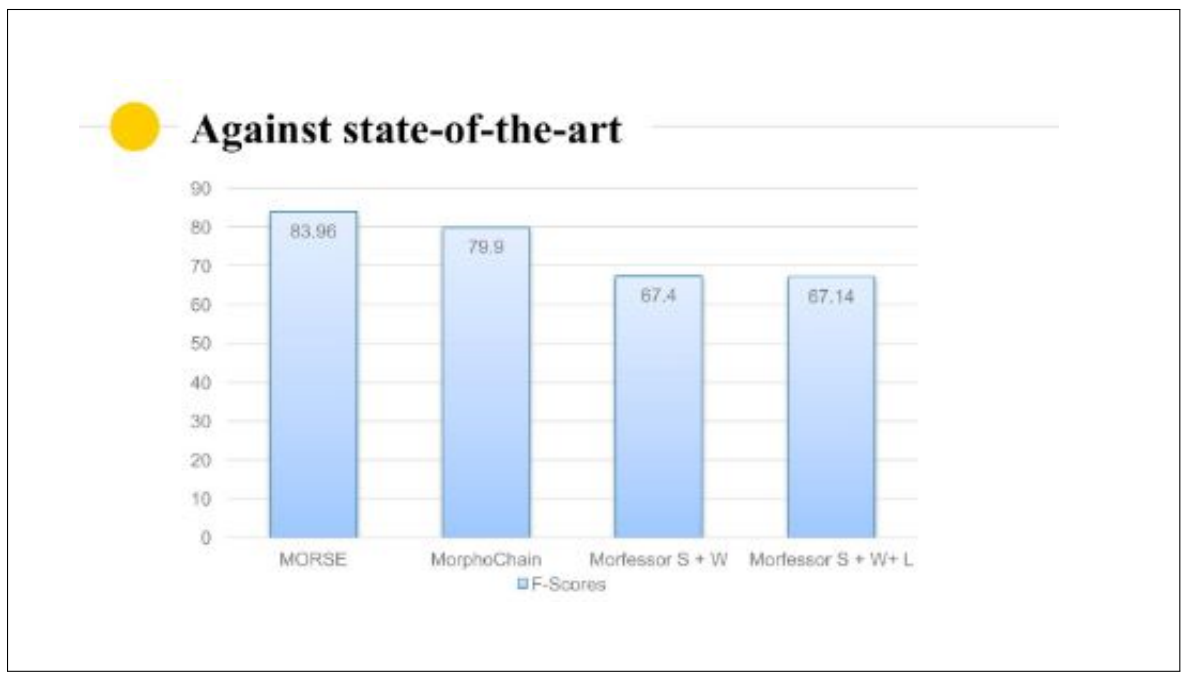

Figure 16: An Example from SciDoc2DiagramBench-Gold where the Intent is "Create a Bar chart showing the comparison of F1 score of MORSE, Morfessor S+W, Morfessor S+W+L and MorphoChain against published state-of-the-art results" and the source paper is (Sakakini et al., 2017)

documents, capturing intricate diagram intent from user-specified slides or text sources, making our method applicable for use in research papers and presentations.

While ChartMimic and MMCode focus on generating code that addresses visual understanding or mathematical problems, our method generates scientific diagrams for seamless integration into academic presentations. This involves translating user intent into structured diagrams that comply with scientific standards. To the best of our knowledge, none of the previous approaches have specifically addressed the challenge of multimodal code generation for scientific diagrams with the focus on preserving layout, completeness, and faithfulness in a highly structured and accurate form. This gap is what SciDoc2DiagramBench fills, making it a novel contribution.

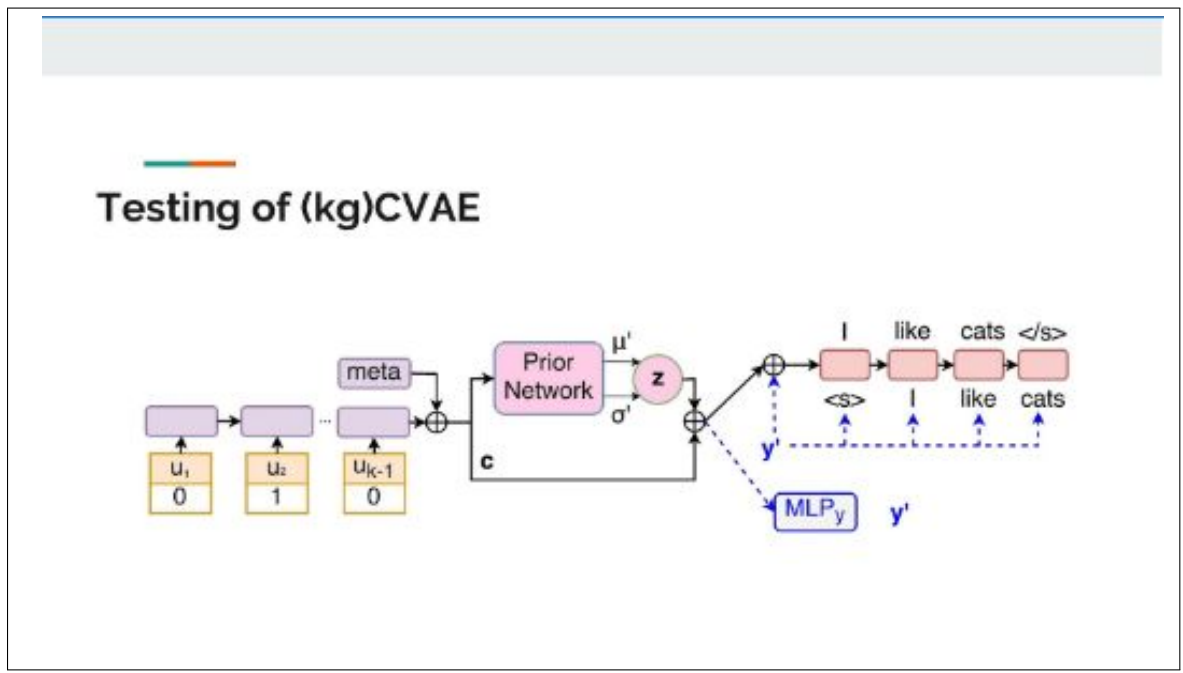

Figure 17: An Example from SciDoc2DiagramBench-Gold where the Intent is "Create a Bar chart showing the comparison of F1 score of MORSE, Morfessor S+W, Morfessor S+W+L and MorphoChain against published state-of-the-art results" and the source paper is (Zhao et al., 2017)

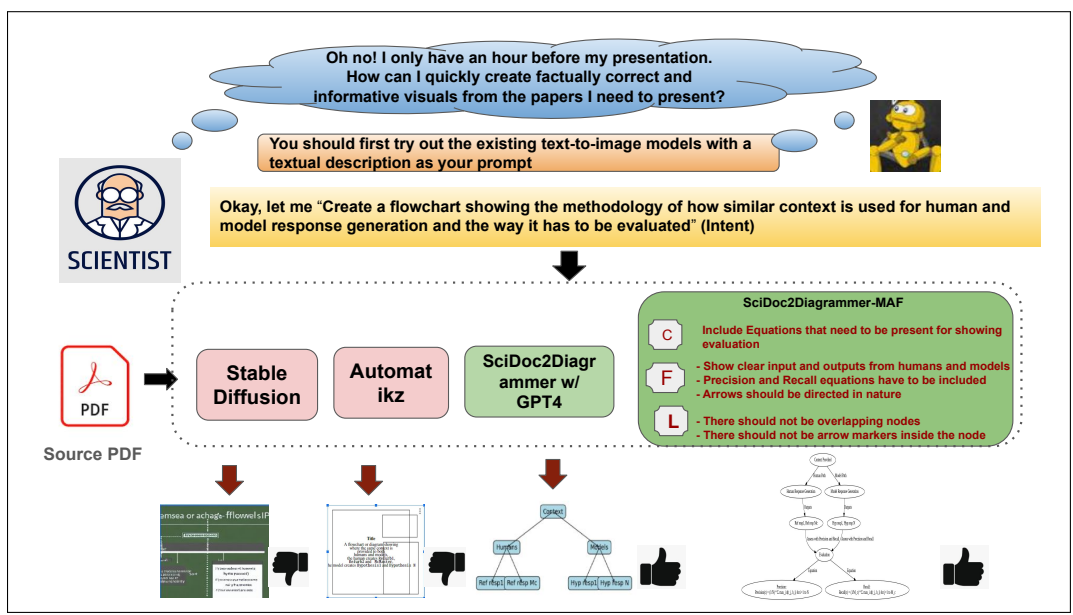

Figure 18: A busy scientist is inclined to automate the creation of diagrams, where our refinement-based approach in SciDocDiagrammarMAF yields a complete and faithful diagram.

In addition, human-computer interaction is an emerging topic as artificial intelligence (AI) advances in diverse tasks. However, AI still faces challenges such as gender and race discrimination (Nghiem et al., 2024; An et al., 2024), producing unfaithful information through hallucinations (Huang et al., 2023), privacy breaches (Yao et al., 2024; Zhou et al., 2024), etc. Fully rely-ing on LLMs for diagram generation may overlook specific user intentions and use cases. Moreover, users learn their mental models of the data and refine their criteria and results through the interaction with AI (Shankar et al., 2024; Li et al., 2024b). Future work could focus on interactive diagram generation, allowing users to iteratively engage with AI to refine and generate diagrams that best suit their specific use cases.

| Strategy | Description |
| --- | --- |
| **DALLE3 with GPT4-o** (Rombach et al., 2021) | Uses GPT4-o output for diagram planning and diagram rendering using DALLE3 in SciDoc2Diagrammer. |
| **Automatikz** (Belouadi et al., 2023) | Uses GPT4-o output for diagram planning and diagram rendering using Clima-13B (https://huggingface.co/nllg/tikz-clima-13b/) in SciDoc2Diagrammer. |
| **Zero-shot SciDoc2Diagrammer w/ Phi-3** (Abdin et al., 2024) | Implements Phi-3 (https://huggingface.co/microsoft/Phi-3-vision-128k-instruct) for diagram planning and GPT4-o for diagram rendering with code generation. |
| **Few-shot SciDoc2Diagrammer w/ Phi-3** (Abdin et al., 2024; Rozière et al., 2024) | Like Zero-shot but uses 3 examples for guidance. |
| **Zero-shot SciDoc2Diagrammer w/ Mistral** (https://huggingface.co/mistralai/Mistral-7B-Instruct-v0.2) (Jiang et al., 2023) | Uses Mistral for diagram planning and GPT4-o for diagram rendering with code generation. |
| **Few-shot SciDoc2Diagrammer w/ Mistral** (Jiang et al., 2023; Rozière et al., 2024) | Extends Zero-shot by using 3 example inputs. |
| **Zero-shot SciDoc2Diagrammer w/ GPT4-o** (OpenAI, 2023) | Utilizes GPT4-o for diagram planning and diagram rendering in SciDoc2Diagrammer. |
| **Few-shot SciDoc2Diagrammer w/ GPT4-o** (OpenAI, 2023) | Adds 3 exemplars to GPT4-o based generation process. |

Table 8: Description of Base and Refiner LMs

| Models | CS | BS | C | F | L |
| --- | --- | --- | --- | --- | --- |
| SciDoc2Diagrammer | 0.51 | 0.61 | 3.8 | 4.3 | 3.8 |
| - w/o QG/QA | 0.34 | 0.49 | 2.8 | 4.0 | 3.7 |
| - w/o Summarization | 0.43 | 0.53 | 3.2 | 4.0 | 3.8 |

Table 9: Influence of components in SciDoc2Diagrammer evaluated on **SciDoc2DiagramBench-Gold** using metrics such as CLIPScore (CS), BERTScore (BS), Completeness (C), Faithfulness (F), and Layout (L) using GPT4-Evaluation, showing that QA/QG and Information summarization contributes to the best performance across all metrics.

| Error Taxonomy in Generated Diagrams | Definition of Error Categories | Prevalence in the Type of Diagrams | Error Propagation from which step |
|---|---|---|---|
| Missing Step or Node | Omission of a crucial process step or component | Tables Summary, Flowcharts | IE to Code Generation |
| Missing Arrow | Absence of an arrow that indicates flow between elements | Flowcharts | IE to Code Generation |
| Misleading Information | Incorrect details that contradict the intended data/narrative | Flowcharts, Summary Tables | IE to Code Generation |
| Intent Mismatch | Diagram does not align with user-defined intent | All Types | Intent Classification |
| Incorrect Order of Nodes | Nodes appear in an incorrect sequence | Flowcharts | IE to Code Generation |
| Wrong Numerical Value/Trends | Numerical inaccuracies or incorrect trends | Results/plots | Data Extraction, IE to Code Generation |
| Wrong Reasoning | Logical errors leading to incorrect conclusions | Results/plots | IE to Code Generation |
| Wrong Axes Labels | Axes mislabeling, including incorrect titles or units | Results/plots | Data Extraction, IE to Code Generation |
| Incomplete Reasoning | Diagrams that leave out critical aspects of analysis | Results/plots, Summary Tables | IE to Code Generation |
| Unreadable Fonts | Fonts that are difficult to read, diminishing accessibility | Results/plots, Flowcharts | IE to Code Generation |
| Overlapping Legends or Title of the Text | Legends or titles overlap with other elements, obscuring information | Results/plots, Flowcharts | IE to Code Generation |
| Incorrect Bounding Box | Poorly sized/misplaced bounding boxes | Explanation of Figures | IE to Code Generation |

Table 10: Comprehensive Error Taxonomy which categorizes the common errors identified on a randomly selected set of 50 generated diagrams from scientific documents from the test set of SciDoc2DiagramBench w/ the best performing GPT4-o which had GPT4-Eval score less than 3.5 detailing the nature of each error.

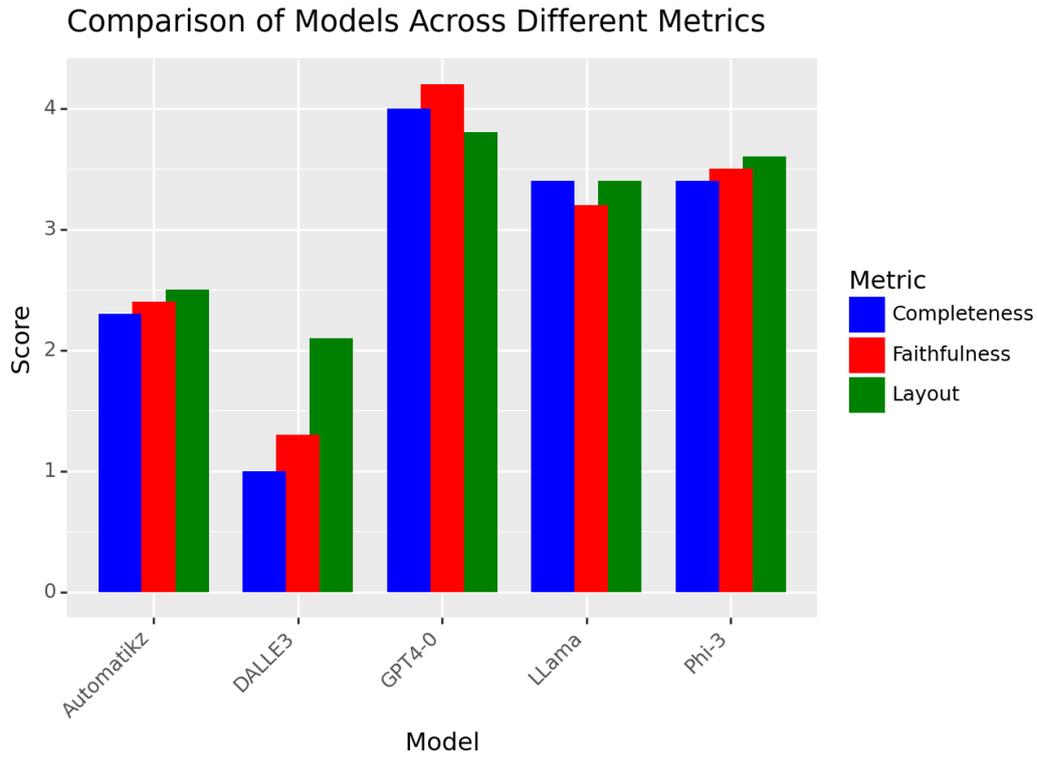

Figure 19: Humans Rated on the diagrams generated by the SciDoc2Diagrammer on three evaluation criteria. We observe that GPT4-o is clearly the winner model.

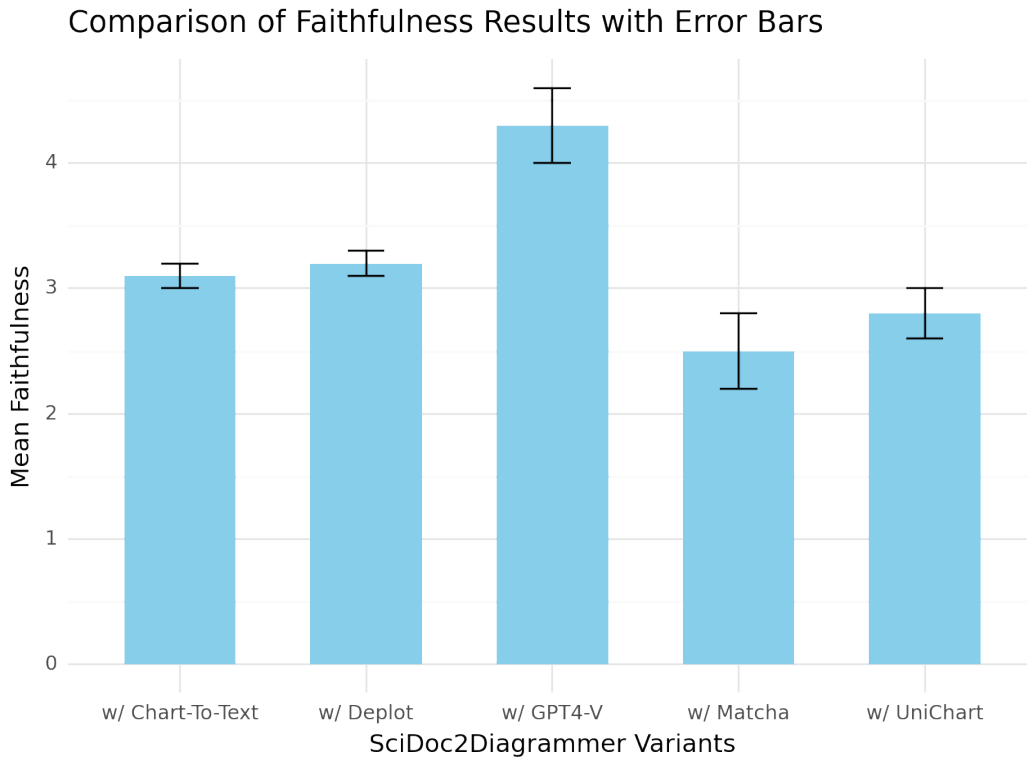

Figure 20: Faithfulness evaluated on a sample of 30 diagrams by two annotators (On a Likert Scale 1-5) when shown the original figure and the generated figure.

| Critic | Base LM and Variants | C | F | L |
|---|---|---|---|---|
| GPT4-V | DALLE3 | 1.5 | 1.0 | 2.0 |
|  | Automatikz | 2.6 | 3.2 | 3.5 |
|  | Phi-3 | 3.4 | 4.3 | 3.5 |
|  | GPT4o | 3.8 | 4.3 | 3.8 |
|  | GPT4o + MAF | 4.6 | 4.7 | 4.6 |
| Human | DALLE3 | 1 | 1.2 | 2.1 |
|  | Automatikz | 2.8 | 2.4 | 3.2 |
|  | Phi-3 | 3.5 | 3.4 | 3.1 |
|  | GPT4o | 3.6 | 4.0 | 3.4 |
|  | GPT4o + MAF | 4.2 | 4.4 | 3.6 |

Table 11: Average Completeness (C), Faithfulness (F), and Layout (L) comparing the performance of models, evaluated by both GPT4-Vision and human evaluators.

| Refinement | Completeness | Faithfulness | Layout |
|---|---|---|---|
| w/ MAF | 81% | 75% | 56% |
| w/ SR | 51% | 54% | 51% |

Table 12: LLM-Human Agreement Scores for Refined vs. Non-Refined Images

|  | Extrapolated-Flowchart | | | Extrapolated-Results | | | Extrapolated-Architecture | | | Extrapolated-Summary | | |
|---|---|---|---|---|---|---|---|---|---|---|---|---|
| Models (IV) | R | BS | CS | R | BS | CS | R | BS | CS | R | BS | CS |
| DALLE-3 | 0.12 | 0.45 | 0.23 | 0.11 | 0.34 | 0.21 | 0.25 | 0.45 | 0.25 | 0.04 | 0.10 | 0.18 |
| Automatikz | 0.20 | 0.48 | 0.25 | 0.20 | 0.50 | 0.29 | 0.24 | 0.45 | 0.28 | 0.27 | 0.56 | 0.45 |
| SciDoc2Diagrammer | | | | | | | | | | | | |
| w/ LLama2 | 0.28 | 0.58 | 0.40 | 0.28 | 0.39 | 0.28 | 0.24 | 0.48 | 0.45 | 0.45 | 0.58 | 0.54 |
| w/ Phi-3 | 0.26 | 0.66 | 0.52 | 0.32 | 0.47 | 0.38 | 0.28 | 0.41 | 0.38 | 0.40 | 0.67 | 0.56 |
| w/ Mistral | 0.26 | 0.64 | 0.45 | 0.39 | 0.53 | 0.34 | 0.29 | 0.48 | 0.42 | 0.39 | 0.59 | 0.62 |
| w/ GPT4-o | 0.28 | 0.67 | 0.43 | 0.40 | 0.56 | 0.38 | 0.32 | 0.57 | 0.45 | 0.45 | 0.67 | 0.62 |

Table 13: Automatic evaluation of models on various diagrams on **SciDoc2DiagramBench-Gold** using ROUGE (R), BERTScore (BS), and CLIPScore (CS). The table highlights the comparative performance of DALLE-3, Automatikz, and several versions of SciDoc2Diagrammer (Few-shot Strategy) across all categories of diagrams. SciDoc2Diagrammer with GPT4-o consistently shows the best performance across most categories, indicated by the green cells, while DALLE-3 generally underperforms, as shown by the red cells. We observe that GPT4-o is clearly the winner model.

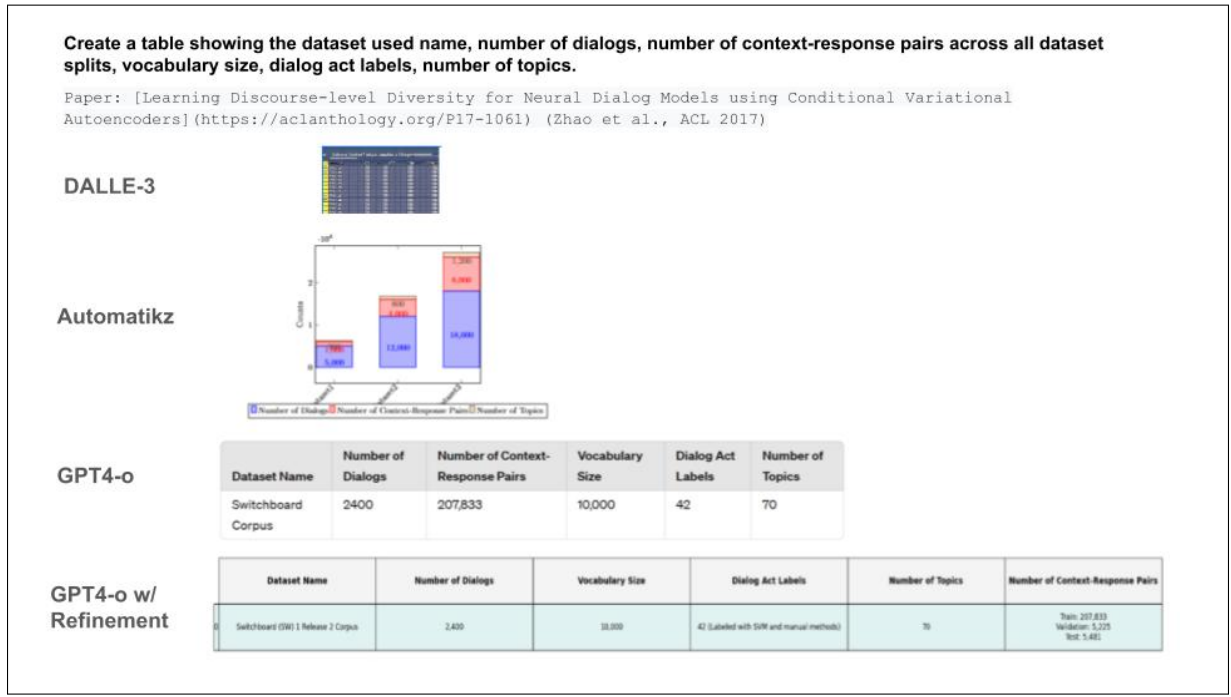

Figure 21: Qual Examples

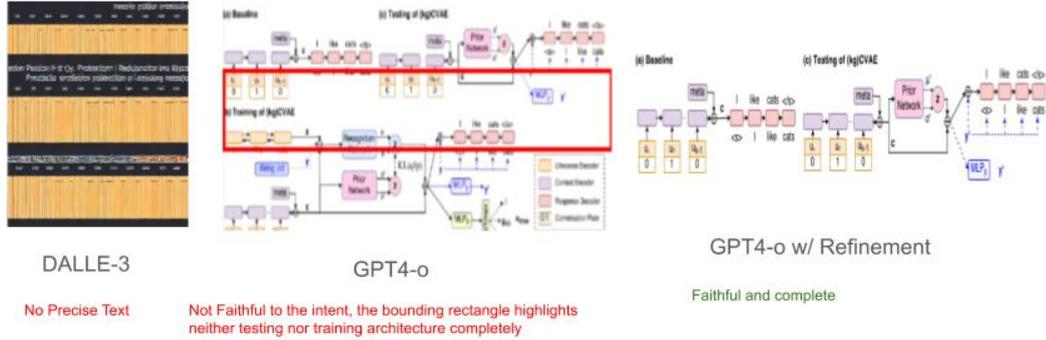

Figure 22: Qual Examples

---

**Algorithm 1** Completeness Assessment ($C_{critic}$)

**Require:** $O$: Diagram to be evaluated, $Code$: Associated Code to be Checked
**Require:** $PDF$: Document source
**Require:** $I$: Defined intent for diagram creation
**Ensure:** $C$: Completeness score of $O$ and $C_{Feedback}$ to improve the score
1: $S = []$
2: Decompose $I$ and generate a set of questions $Q$ based on the decomposed intent
3: **for** each question $q_i$ in $Q$ **do**
4:     $a_i \leftarrow$ extract$(q_i, O)$     ▷ Get answer determining completeness of $O$
5:     $s_i \leftarrow [a_i$ is adequately represented in $O]$     ▷ Assign Score from 1-5
6:     Append $s_i$ to score list $S$
7: $C_{Score} \leftarrow \frac{1}{|Q|} \sum_{i=1}^{|Q|} s_i$     ▷ Calculate average score
8: $C_{Feedback} \leftarrow$ textual feedback to improve $C$
9: **return** $C_{Score}$ and $C_{Feedback}$

---

**Require:** $D$: Diagram to be evaluated
**Require:** $R$: Design rules for layout and aesthetics
**Ensure:** $S$: Aesthetics score of diagram $D$
**Ensure:** $F$: Feedback for improving $D$
1: **Function** EvaluateLayout($D$, $R$)
2: $S \leftarrow 0$
3: $Feedback \leftarrow []$
4: **for** each rule $r$ in $R$ **do**
5:     $result, comment \leftarrow$ APPLYRULE$(r, D)$
6:     $S \leftarrow S + result$
7:     $Feedback.append(comment)$
8: **return** $S, Feedback$

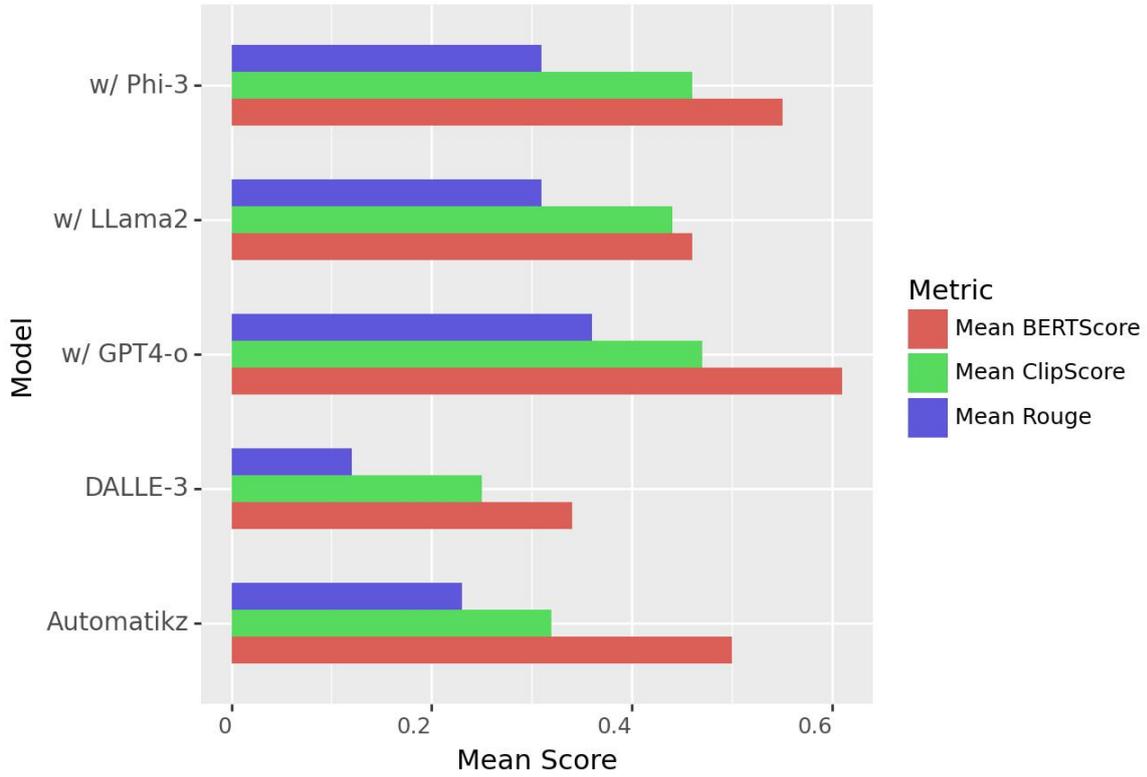

Figure 23: Zero-Shot Evaluation of SciDoc2Diagrammer with various LLMs using Automatic Evaluation Metrics. We observe that GPT4-o is clearly the winner model.

---

**Algorithm 2** Faithfulness Assessment ($F_{critic}$)

---
**Require:** $O$: Diagram to be evaluated and corresponding code $Code$
**Require:** $PDF$: Document source
**Require:** $I$: Defined intent for diagram creation
**Ensure:** $F$: Faithfulness score of $O$ and $F_{Feedback}$ to improve the score
1: Using $O$ and $Code$, generate questions that would validate the correctness of $O$
2: $S = []$
3: **for** each question $q_i$ in $Q$ **do**
4: $\quad a_i^{pdf} \leftarrow \text{retrieve}(q_i, PDF)$ ▷ Retrieve answer from PDF
5: $\quad a_i^{image} \leftarrow \text{extract}(q_i, O)$ ▷ Retrieve answer from diagram $O$
6: $\quad s_i \leftarrow [a_i^{image} == a_i^{pdf}]$ ▷ ▷ Assign Score from 1-5
7: $\quad$ Append $s_i$ to score list $S$
8: $F_{\text{score}} \leftarrow \frac{1}{|Q|} \sum_{i=1}^{|Q|} s_i$ ▷ Calculate average $score$
9: $F_{Feedback} \leftarrow$ textual feedback to improve $C$
10: **return** $F_{Score}$ and $F_{Feedback}$
---

**Algorithm 3** Sequential Refinement of Diagrams
___
**Require:** Diagram $D$, Source PDF $P$
**Ensure:** Refined Diagram $D_{\text{ref}}$
___
1: **for** type in {'Completeness', 'Faithfulness', 'Layout'} **do**
2:     maxIterations $\leftarrow$ 3
3:     threshold $\leftarrow$ 4.5
4:     iteration $\leftarrow$ 0
5:     score $\leftarrow$ 0
6:     **while** iteration $\leq$ maxIterations and score $<$ threshold **do**
7:         feedback $\leftarrow$ Evaluate($D$, $P$, type)       ▷ Evaluation based on type
8:         **if** Score(feedback) $>$ threshold **then**
9:             **break**       ▷ Exit loop if score exceeds the threshold
10:        $D \leftarrow$ RefineDiagram($D$, feedback)       ▷ Refine diagram based on feedback
11:        score $\leftarrow$ Score(feedback)       ▷ Update score based on feedback
12:        iteration $\leftarrow$ iteration + 1
13: **return** $D$
___

**Algorithm 4** Summarization-Based Refinement
___
**Require:** Diagram $D$, Source PDF $P$
**Ensure:** Refined Diagram $D_{ref}$
___
1: $iteration \leftarrow 0$, $maxIterations \leftarrow 3$, $threshold \leftarrow 4.5$, $score \leftarrow 0$
2: **while** $iteration < maxIterations$ OR $score < threshold$ **do**
3:     $C_{Score}, C_{feedback} \leftarrow C_{critic}(D, P)$
4:     $F_{Score}, F_{feedback} \leftarrow F_{critic}(D, P)$
5:     $L_{Score}, L_{feedback} \leftarrow L_{critic}(D)$
6:     Initialize $allFeedback \leftarrow \{\}$
7:     **if** Score($C_{Score}$) $< threshold$ **then**
8:         $allFeedback \leftarrow allFeedback \cup \{C_{feedback}\}$
9:     **if** Score($F_{Score}$) $< threshold$ **then**
10:        $allFeedback \leftarrow allFeedback \cup \{F_{feedback}\}$
11:     **if** Score($L_{Score}$) $< threshold$ **then**
12:        $allFeedback \leftarrow allFeedback \cup \{L_{feedback}\}$
13:     $D \leftarrow RefineDiagram(D, allFeedback)$
14:     $score \leftarrow \text{Min}(C_{Score}, F_{Score}, L_{Score})$
15:     $iteration \leftarrow iteration + 1$
16:     $D_{ref} \leftarrow D$
17: **return** $D_{ref}$

| |
|---|
| **Prompt template for Intent Generation** |
| **[1]** You will be given the title, abstract and Summarized section content of the Paper, figure and table captions in the paper, and you will have to come up with the intent of possible type of a diagram that is not there in the paper, but will be useful to create. |
| **[2]** Come up with an intent which will require you to process mathematical, logical reasoning on top of the extracted data from the paper, and come up with such intent that will require you to blend text and visuals from the paper to come up with the final diagram. |
| **[3]** The intent should be clear, comprehensive and any human can create a diagram from the paper content based on the intent. |
| SOME IN-CONTEXT EXAMPLES |
| Title: |
| Abstract : |
| Section Content : |
| Table Captions: |
| Image Captions: |
| Your Intent: |

Table 14: The prompt template for Intent Generation.

| |
|---|
| **Prompt template for Intent Classification** |
| **[1]** You will be provided with an intent to task is to understand the intent of the diagram creation and classify into one of the following labels: |
| 1) Extrapolated-Flowchart: If the intent is related to creating flowchart |
| 2) Extrapolated-Summary: If the intent is related to summarization of related work, contributions, methods, datasets or hyperparameter or experimental details mentioned in the paper |
| 3) Extrapolated-Architecture: If the intent is related to modifying an architecture or existing image in the paper |
| 4) Extrapolated-Results: If the intent is related to generate plots or visualize results |
| SOME IN-CONTEXT EXAMPLES |
| Intent: |
| Label : |

Table 15: The prompt template for Intent Classification.

| |
|---|
| **Prompt template for Question Generation** |
| **[1]** Your intent of coming up with the diagram creation is provided below. Generate clarification questions based on the intent what information needs to be extracted so that you can generate the diagram. |
| SOME IN-CONTEXT EXAMPLES |
| Intent: |
| Clarification Questions: |

Table 16: The prompt template for Question Generation.

| |
|---|
| **Prompt template for Data Extraction From Figures** |
| **[1]** Extract Raw data in the form of markdown from the image. |
| Image: |
| Extracted Markdown: |

Table 17: The prompt template for Data Extraction From Figures

| **Prompt template for Answer Extraction** |
|---|
| **[1]** Your intent of diagram creation is presented along with the section text or image data. For each of the question, if the section text or image data is relevant, extract answer for the questions, if not relevant then, say 'NA'. Make sure that you do not extract information that is not present in the source code or image. |
| **[2]** Format your output as a list of JSON objects (Question/Answer pairs) where the keys are your questions and answers are the values. SOME IN-CONTEXT EXAMPLES intent: |
| Section/Image : |
| Questions : |
| Question/Answer Pairs: |

Table 18: The prompt template for Answer Extraction.

| **Prompt template for Code Generation** |
|---|
| **[1]** Generate a code in python where the intent of the diagram is provided to you, along with the intent type. The information to be presented is also in front of you. You should use the information to display the content. |
| **[2]** If the intent is about creating a flowchart, it has to be in graphviz. If the intent is about creating plots/line charts/graphs, it has to be clear and legible, ideally in plotnine. If the intent is to create or highlight a portion of image, you should use the pillow library to include bounding box or textual explanation. Also show that if you want to create a summary, it should be in a good layout with proper table header and fonts. |
| SOME IN-CONTEXT EXAMPLES |
| Intent: |
| Intent Type : |
| Question-Answer pairs: |
| Generated Code: |

Table 19: The prompt template for Code Generation.

| **Prompt template for Line 2 in Algorithm 1** |
|---|
| **[1]** You are provided with the intent of diagram creation. Decompose the intent and ask questions such that the answers to those questions will determine completeness of information. |
| SOME IN-CONTEXT EXAMPLES |
| Intent: |
| Questions: |

Table 20: Prompt template for Line 2 in Algorithm 1

| **Prompt template for Answer Extraction of the Question from Image/Associated Code** |
|---|
| **[1]** You are provided with the intent of diagram creation, image and the corresponding code that generated the image as input. Your goal is to extract answer of the question from the image or the code. Make sure that you do not extract information that is not present in the source code or image. |
| Intent: |
| Image: |
| Question: |
| Corresponding Code: |
| Answer of the Question: |

Table 21: Prompt template for Answer Extraction of the Question from Image/Associated Code

| **Prompt template for Refinement Step** |
|---|
| [1] You are provided with diagram and the associated code that generated it. Based on criteria name, the image has received a score of the given score out of 5, and a feedback is generated to improve. Refine the code to incorporate the following feedback.<br>Image:<br>criteria name:<br>Corresponding Code:<br>Score:<br>Feedback :<br>Refined Code: |

Table 22: Prompt template for Refining generated Image/Associated Code

| **Prompt template for Completeness Critic Evaluation** |
|---|
| [1] You are provided with the intent of the diagram creation, answers from PDF and the answers obtained from the Image or the code. You have to determine if the answer from the PDF/intent is completely present in the answer from image/code. Please provide a Completeness score from 1-5, where 1 denotes there is minimal Completeness and 5 when there is an exact match with the PDF Content. If your score is less than 4.5, generate feedback on what is needed to improve the Completeness.<br>Image:<br>Corresponding Code:<br>Intent:<br>Answer from the PDF:<br>Answer from the Diagram/Code:<br>Completeness Score:<br>Feedback : |

Table 23: Prompt template for Completeness Critic Evaluation

| **Prompt template for Faithfulness Critic Evaluation** |
|---|
| [1] You are provided with the intent of the diagram creation, answers from PDF and the answers obtained from the Image or the code. You have to determine if the answer from the Image or the code is faithful or true with respect to answer from the PDF. Please provide a Faithfulness score from 1-5, where 1 denotes there is minimal faithfulness and 5 when there is an exact match with the PDF Content. If your score is less than 4.5, generate feedback on what is needed to improve the faithfulness.<br>Image:<br>Corresponding Code:<br>Intent:<br>Answer from the PDF:<br>Answer from the Diagram/Code:<br>Faithfulness Score:<br>Feedback : |

Table 24: Prompt template for Faithfulness Critic Evaluation

| |
|---|
| **Prompt template for Layout Critic Evaluation** |
| [1] You are provided with the intent of the diagram creation and the image, source PDF. |
| [2] You have to determine how much readable, comprehensible, precise in terms of the look-and-feel of the diagram. By precision, it means that all the necessary scientific information conveyed in the image can be easily deciphered. Please provide a score from 1-5, where 1 denotes there is minimal satisfaction on the look-and-feel aspect and 5 when there is nothing to complain about the look-and-feel aspect. If your score is less than 4.5, generate feedback on what is needed to improve the look-and-feel. |
| Image: |
| Intent: |
| Layout Score: |
| Feedback : |

Table 25: Prompt template for Layout Critic Evaluation

| |
|---|
| **Prompt template for Self-Refine Evaluation** |
| [1] You are provided with the intent of the diagram creation and the image. |
| [2] There might be some problem inside the image. Please generate feedback if you feel that there has been any inconsistency. |
| Image: |
| Intent: |
| One-step Score: |
| Feedback : |

Table 26: Prompt template for Self-Refine Evaluation